\pdfoutput=1
\documentclass[11pt]{article}

\usepackage[]{acl}

\usepackage[table]{xcolor}
\newcommand{\best}[1]{\cellcolor{Grey!10}\textbf{#1}} %
\usepackage{xcolor}
\definecolor{Grey}{rgb}{0.5,0.5,0.5}

\usepackage{times}
\usepackage{makecell}   % add this in preamble if not already there
\usepackage{longtable}
\usepackage{latexsym}
\usepackage{booktabs}
\usepackage{tabularx}
\usepackage{hyperref}
\usepackage{adjustbox}
\usepackage{multirow}
\usepackage{caption} % in preamble
\usepackage{svg}
\usepackage{comment}
\usepackage{listings}
\usepackage[table]{xcolor}
\usepackage[most]{tcolorbox}
\usepackage{tabularx}
\usepackage{enumitem}
\usepackage{rotating}
\usepackage{float}
\usepackage{graphicx}

\usepackage[T1]{fontenc}
\usepackage[utf8]{inputenc}

\usepackage{microtype}

\usepackage{inconsolata}

\usepackage{graphicx}
\title{\textit{BaatCheet:} A Multilingual Corpus for Dialogue Translation in \\ Indian Languages}

\author{
    \textbf{Priyanka Dasari}\thanks{Authors contributed equally.},
    \textbf{Yuvrajsinh D. Bodana}\footnotemark[1],
    \textbf{Vandan Mujadia}, \\
    \textbf{Arafat Ahsan},
    \textbf{Dipti Misra Sharma},
    \textbf{Parameswari Krishnamurthy} \\[4pt]
    \textit{Language Technologies Research Centre, IIIT Hyderabad, India} \\[4pt]
    \texttt{\{dasari.priyanka,yuvrajsinh.bodana\}@research.iiit.ac.in}, \\
    \texttt{vmujadia@gmail.com},
    \texttt{\{arafat.ahsan,dipti,param.krishna\}@iiit.ac.in}
}

\begin{document}
\maketitle
\begin{abstract}
 
Existing translation models are typically trained on sentence-level and formal text, limiting their ability to capture everyday conversational dialogue phenomena such as informality, speaker interaction, and discourse 
coherence. 
Most existing Indic translation resources and evaluation benchmarks focus on
sentence-level or formal text, making it difficult to assess translation quality of the dialogue phenomena.
In this work, we introduce \textit{BaatCheet}, a multilingual dialogue corpus named after 
the Hindi term for \textit{conversation} or \textit{chitchat}, 
containing approximately 49,000 dialogues for dialogue translation across five 
translation directions. 
We fine-tune five open-source LLMs across seven training data configurations 
and find that fine-tuning yields substantial gains over zero- and few-shot 
baselines. 
To comprehensively evaluate dialogue translation quality, 
we employ multiple evaluation strategies, including automatic metrics, LLM-as-judge, and human assessments using an SQM-guided Direct Assessment (DA) Protocol. The dataset is available here\footnote{https://huggingface.co/datasets/HimangY/BaatCheet-Corpus}.
% \footnotetext{\textsuperscript{*}These authors contributed equally.}
\end{abstract}

\section{Introduction}

% Translating informal spoken dialogue presents challenges that are qualitatively different from those of standard Machine Translation (MT). While MT research has long been driven by formal translation tasks~\cite{koehn-2005-europarl}, translating informal text is substantially harder~\cite{farajian-etal-2016-measuring}, and dialogue in particular remains a largely unaddressed frontier~\cite{bawden-etal-2020-diabla}. Informal dialogue is characterized by code-mixing~\cite{sitaram-etal-2020-survey, bali-etal-2014-borrowing},  ellipsis, and colloquial expressions that reflect the spontaneity of natural speech~\cite{gumperz1982language};  sentence-level MT systems routinely fail on such phenomena, producing translations that are locally fluent but globally inconsistent~\cite{voita-etal-2019-good}. 

\begin{figure}[t!]
\centering
\includegraphics[height=0.25\textheight]{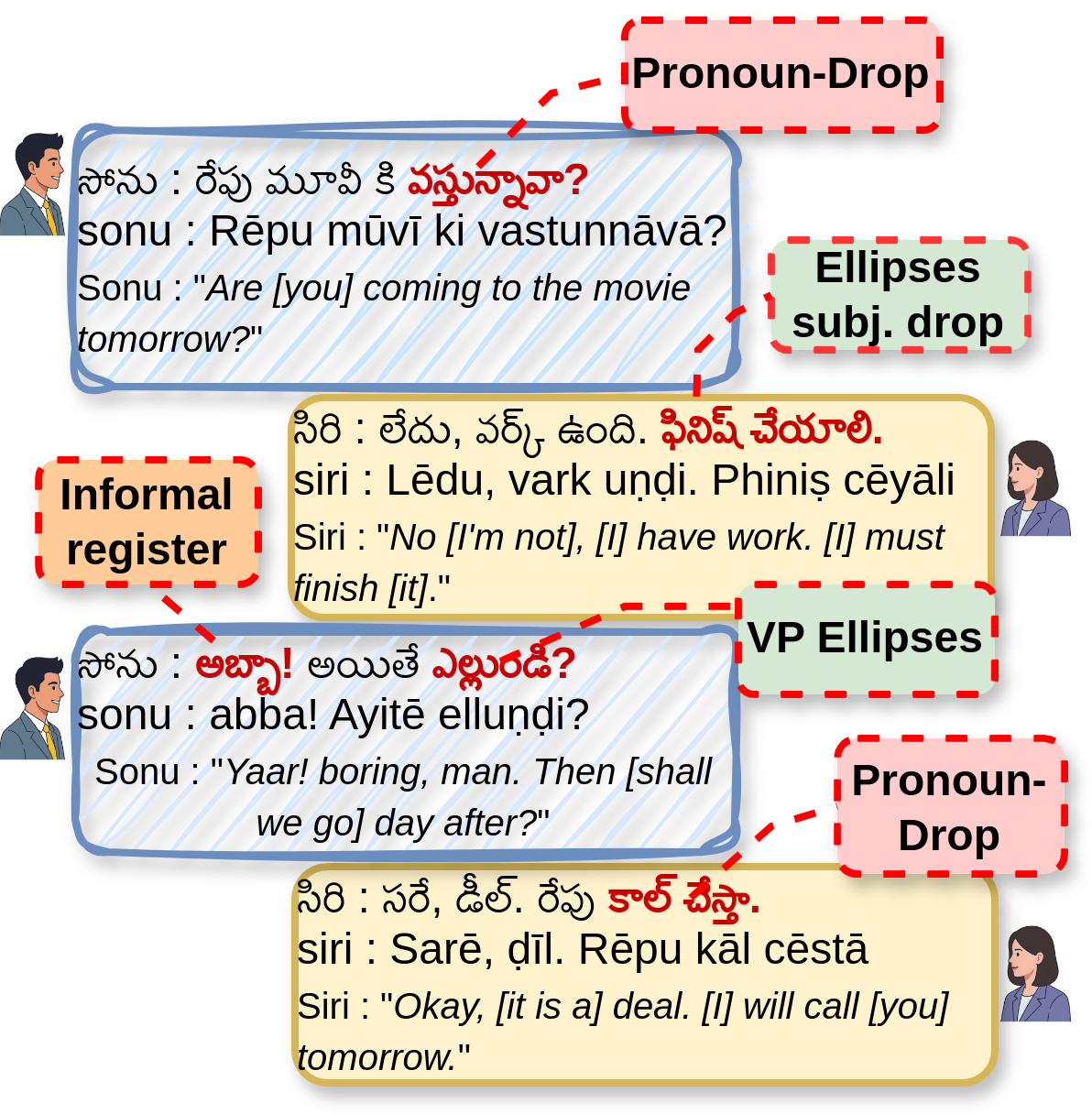}
\caption{\small  Highlighting Linguistic phenomena in informal Telugu dialogue. Dashed boxes highlight examples of pro-drop, ellipses, and register-specific informal expressions in a sample conversation.}
\label{fig:example_dia}
\end{figure}

\textit{Dialogue} is a fundamental mode of human communication through which 
individuals express emotions, intentions, and maintain social relationships~\cite{gumperz1982language,halliday1987spoken}.
In the sociolinguistic context of the Indian subcontinent, these interactions are predominantly multi-turn and informal, often involving dynamic code-mixing between regional and global languages. Such dialogues diverge from formal written text, which exhibits fragmented syntax, elliptical structures, and heavy use of colloquial expressions that reflect the spontaneity of natural speech. 
With the growing need for cross-lingual communication in multilingual societies like India, machine translation (MT) has become pivotal in bridging language barriers. Most existing MT systems are primarily trained on formal, sentence-level parallel corpora derived from news and governance domains \cite{koehn-2005-europarl, mujadia2025bhashaversetranslationecosystem}. Such systems often struggle to preserve the pragmatic and contextual nuances inherent in natural conversation \cite{sung-etal-2024-context} and this gap persists even for LLM-based translation systems~\cite{choudhary-etal-2025-exploring}, a trend our own baseline evaluation reaffirms for informal Indic dialogue (\S\ref{mt_selection}).

To illustrate these challenges, we present a sample dialogue in Telugu, a major Dravidian language\footnote{Throughout this paper we use \textit{Indic} as an umbrella term for the languages of the Indian subcontinent covered in our corpus, spanning both the Indo-Aryan (e.g., Hindi, Gujarati) and Dravidian (e.g., Tamil, Telugu) language families; we use \textit{Dravidian} specifically when referring to that family.}, in Figure~\ref{fig:example_dia}. Unlike the formal datasets commonly used in MT research, this spontaneous conversation is replete with code-mix, pro-drop (where both subjects and objects are null) and contextual ellipsis, where utterances like `No' (lēdu) and `Then day after?' (ayitē elluṇḍi?) carry a high semantic load that is only recoverable through multi-turn reasoning.

For example, a major challenge in dialogue translation is the encoding of social relationships through honorific pronouns. Unlike English, which uses a single second-person pronoun \textit{you} across all social contexts, Indian languages grammaticalize social distance and respect, directly distinguishing formal and informal pronouns.  These distinctions carry pragmatic weight that MT systems routinely collapse~\cite{hingrajiya-etal-2026-indicdisco}.
This limitation is highly consequential in high-stakes applications (e.g., healthcare, customer support, and travel assistance), where misinterpreting register or speaker intent can degrade communication. To evaluate these phenomena, sentence-level parallel test sets exist for Indic 
languages~\cite{nllbteam2022languageleftbehindscaling}, but no dialogue-level test set is available for evaluating informal dialogue translation. 
% This absence makes it difficult to even measure how well current systems whether general-purpose LLMs or dedicated Indic MT engines handle informal style, code-mixing, and discourse coherence in dialogue, let alone to train models specifically for this setting. 

\begin{figure*}[ht!]
\centering
\includegraphics[height=0.14\textheight]{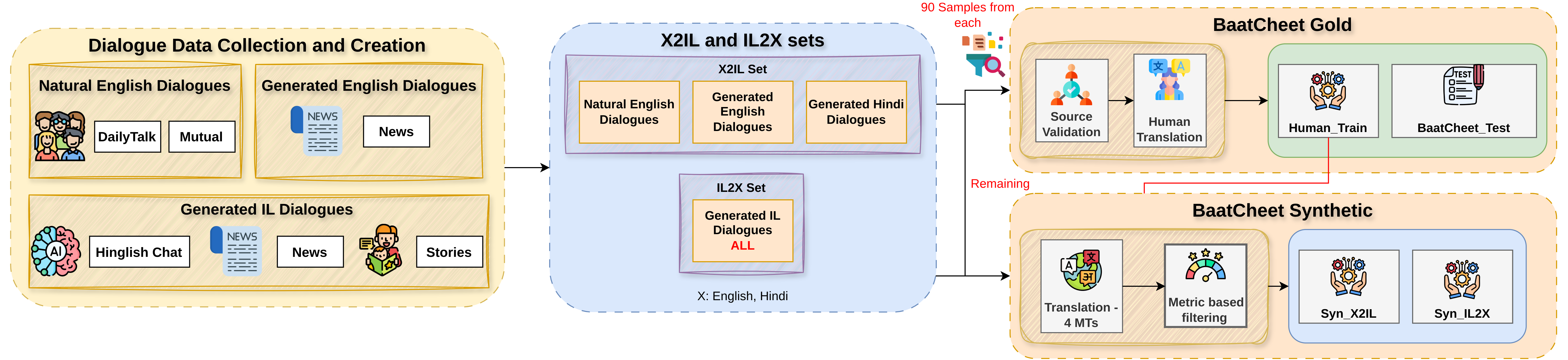}
\caption{\small A flowchart of the BaatCheet Corpora preparation process for training and evaluation. The Metric-based methodology for selecting high-quality synthetic is detailed in Section~\ref{Train-Data}.}
\label{fig:data_pipeline}
\end{figure*}

We address this gap by introducing \textbf{BaatCheet}, named after the Hindi term for \textit{conversation} or \textit{chitchat}, a multilingual corpus for 
informal dialogue translation spanning English, Hindi, Tamil, and Telugu. 
 It is built from combining existing dialogue resources and synthetically generated dialogues, as illustrated in Figure~\ref{fig:data_pipeline}. The dataset consists of two primary components: (i) \textbf{BaatCheet\_Gold}, containing gold-standard human translations split into \textit{BaatCheet\_Human\_Train} and a \textit{BaatCheet\_Test}; and (ii) \textbf{BaatCheet\_Synthetic}, comprising two training subsets: \textit{BaatCheet\_Syn\_X2IL}- a set whose monolingual dialogue data was collected and created in source languages and translated into target languages and \textit{BaatCheet\_Syn\_IL2X} whose monolingual dialogue data was collected and created in target languages and those were translated into source languages and column swapped to make it source to target parallel data for fine-tuning purpose.

Using this corpus, we fine-tune five open-source LLMs across seven training 
data configurations spanning synthetic-only, human-only, and hybrid 
compositions to study how data provenance, scale, and quality interact 
with model capacity. To evaluate translation quality beyond automatic metrics, we conduct human and LLM-as-a-judge evaluations using the SQM-guided DA framework \cite{bojar-etal-2017-results, perrella-etal-2024-beyond}.

The primary contributions of this work are as follows:
\begin{enumerate}[noitemsep,topsep=2pt,leftmargin=*]
    \item \textbf{\textit{BaatCheet}}: A multilingual dialogue corpus for English $\rightarrow$ \{Hindi, Telugu, Tamil\} and Hindi $\rightarrow$ \{Telugu, Tamil\} featuring human translations and synthetically generated dialogues.
    \item \textbf{\textit{BaatCheet\_Gold}}: A curated gold-standard testset for evaluating informal dialogue machine translation in Indic languages.
    \item A structured data creation pipeline employing automatic metric-guided quality selection for high-quality synthetic training corpus curation.
    \item A fine-tuning study of five open-source LLMs across multiple data configurations, tailored for informal dialogue translation.
    \item A large-scale human and LLM-as-judge evaluation using SQM-guided DA establishing the reliability of automatic and LLM-based assessment for this task.
\end{enumerate}

Specifically, our work addresses the following key research questions:
\begin{itemize}[noitemsep,topsep=2pt,leftmargin=*]
    \item \textbf{RQ1:} How well do existing open-source LLMs, dedicated MT systems, and proprietary LLMs translate informal dialogues into Indic languages?
    \item \textbf{RQ2:} How do the two synthetic data axes, \textit{BaatCheet\_Syn\_X2IL} and \textit{BaatCheet\_Syn\_IL2X}, differ in their effect on fine-tuned dialogue translation quality? 
    \item \textbf{RQ3:} How well do automatic metrics, and LLM-based judges agree with human assessment of informal dialogue translation?
\end{itemize}

% Early research on dialogue and document-level machine translation focused on incorporating broader discourse context to improve translation coherence. The survey by \cite{maruf2021survey} provides a comprehensive overview of methodologies, training strategies, and evaluation metrics in document-level neural machine translation (NMT).
% \cite{maruf2018contextual} addressed the challenge of bilingual, multi-speaker dialogues by leveraging both source- and target-side conversational history, using datasets such as Europarl v7 \cite{koehn-2005-europarl} and OpenSubtitles2016 \cite{lison2016opensubtitles2016}. Similarly, \cite{vincent2021towards} explored personalized NMT and document-level MT by integrating extra-textual factors such as speaker identity and dialogue cohesion.
\section{Related work}

Early research on document and dialogue-level machine translation focused on incorporating broader discourse history to improve translation coherence~\cite{maruf2021survey}. Traditional approaches leveraged source and target-side conversational history to model multi-speaker dynamics~\cite{maruf2018contextual}, or integrated extra-textual speaker identity cues to improve personalized translation~\cite{vincent2021towards}. The emergence of multilingual Large Language Models (LLMs) has further advanced this domain by enabling zero-shot context handling~\cite{zhang2023prompting} and generating high-quality synthetic parallel text to address low-resource gaps~\cite{jiao2023chatgpt}.

Another line of work focuses on constructing dialogue datasets to support conversational machine translation research. The DailyDialog dataset \cite{li-etal-2017-dailydialog} provides a widely used multi-turn conversational dataset containing human-written everyday dialogues. Subtitle-based resources such as OpenSubtitles \cite{lison2016opensubtitles2016} have also been used extensively for training conversational translation systems.  \cite{wang2016automatic} proposed methods to automatically construct parallel dialogue corpora by aligning subtitle data with monolingual scripts while preserving speaker and discourse information. More recently, \cite{majewska2023cross} introduced an outline-based approach for generating multilingual dialogue datasets, while \cite{bawden2021diabla} developed the \textit{DiaBLa} English–French corpus of informal written dialogues specifically designed, for evaluating context-aware machine translation systems. Additionally, IndicDialogue \cite{arnob2024indicdialogue} provides subtitle-based dialogue corpora across multiple Indic languages.

For Indic languages, benchmarks like FLORES-200~\cite{nllbteam2022languageleftbehindscaling} and the conversational subset IN22-Conv~\cite{gala2023indictrans2highqualityaccessiblemachine} evaluate translation across standard formal registers.
% Large-scale multilingual datasets such as FLORES-200 \cite{nllb2022} provide standardized evaluation benchmarks across hundreds of language directions, including several Indic languages. AI4Bharat introduced the IN22 benchmark, which evaluates translation across 22 Indian languages and includes a conversational subset, IN22-Conv, designed to evaluate translation in conversational domains \cite{gala2023indictrans}. 

However, many of these datasets focus primarily on sentence-level translation or subtitle-style dialogues and often lack the informal conversational structures, and code-mixed characteristics that are common in real-world multilingual conversations across Indian languages. Standard reference-based metrics like BLEU and sentence-level COMET~\cite{rei-etal-2022-comet} often fail to capture dialogue-level coherence and register fidelity. Recent advances introduce context-aware metrics like DOC-COMET~\cite{vernikos2022embarrassingly} alongside LLM-as-a-Judge frameworks for qualitative assessment~\cite{zheng2023judgingllmasajudgemtbenchchatbot}.

% Our \textit{\textbf{BaatCheet}}, an informal dialogue corpus, supported by a rigorous multi-judge evaluation framework comparing automatic metrics, LLM judges, and expert human linguists, directly addresses this research gap. 

% Furthermore, these literature gaps directly motivate our systematic investigation of dialogue translation quality, synthetic data curation, and dialogue-specific metric reliability, which we formalize through our guiding research questions (\textbf{RQ1}--\textbf{RQ5}).

% \section{Building BaatCheet Corpora}\label{BaatCheet}
% \textit{BaatCheet Corpora} is a multilingual dialogue corpus for English, Hindi, Tamil, and Telugu, designed to support training and evaluation of MT systems for informal dialogue. In total, the corpus comprises approximately 49,000 dialogues, 383,000 utterances, and 700,000 sentences. 
% The corpus is constructed from two sources: existing dialogue datasets and synthetically generated dialogues, as illustrated in Figure~\ref{fig:data_pipeline}. Precise dataset split details are reported in the corpus statistics table (Table~\ref{tab:existing_stats}).

\section{Building BaatCheet Corpora} \label{BaatCheet}

\textit{BaatCheet Corpora} is a multilingual dialogue corpus spanning 
English, Hindi, Tamil, and Telugu, designed to support training and 
evaluation of LLMs and MT systems for informal dialogue. 
In total, the corpus comprises approximately 49,000 dialogues, 383,000 utterances, and 690,500 sentences, organized into two major components: 
(i)~\textit{BaatCheet\_Synthetic}, a large-scale machine-translated 
corpus subdivided into \textit{BaatCheet\_Syn\_X2IL} and 
\textit{BaatCheet\_Syn\_IL2X}; and 
(ii)~\textit{BaatCheet\_Gold}, a human-translated set 
subdivided into \textit{BaatCheet\_Human\_Train} and \textit{BaatCheet\_Test}.

% The corpus is built from two complementary components: \textit{BaatCheet\_Gold}, a professionally human-translated set that seeds both evaluation and COMET-based quality filtering, and \textit{BaatCheet\_Synthetic}, a large-scale synthetically curated corpus. Precise statistics across all splits are reported in Table~\ref{tab:existing_stats}.

\paragraph{Data Sources.}
The corpus draws on three categories of dialogue sources spanning all four
languages:
\textbf{(i)~Natural English Dialogues:} \textit{DailyTalk}~\cite{lee2022dailytalk}, and \textit{MuTual}~\cite{mutual} are human-authored two-speaker daily conversational corpora.
\textbf{(ii)~ Generated English Dialogues:} \textit{News-Eng*}\footnote{All * datasources are generated dialogues with GPT-4o-mini}, informal
dialogues generated using GPT-4o-mini from English news articles.
\textbf{(iii)~ Generated IL Dialogues:} \textit{Hinglish-Chat}~\cite{Hinglish-Chat-21M}, an existing LLM-generated dialogue dataset subsequently refined via prompting to reduce
robotic phrasing and improve conversational fluency; and \textit{News-Hin*},
dialogues generated from Hindi news articles, \textit{News-Tam*}, \textit{News-Tel*} and \textit{Story-Tam*}, \textit{Story-Tel*} dialogues generated from Tamil and Telugu news articles and short stories, respectively.
A full description of each data source, including generation prompts and language coverage, is provided in Table~\ref{tab:data_sources_description} (Appendix~\ref{corpora_building}).
\paragraph{Human Validation of Generated Dialogues.}
Large Language Models often generate synthetically fluent text that may suffer from contextual pronoun mismatches, rigid/overly formal phrasing, or awkward code-mixing. To address this, native speakers validated and edited a small sample of LLM-generated dialogues, which comprise both BaatCheet\_Gold source dialogues (Eng-src: News-Eng and Hin-src: Hinglish, News-Hin) and other generated dialogue sources across all languages. 
Across 344 dialogues containing 4,487 turns, 71.80\% of dialogues required at least one human intervention, with 37.24\% of overall individual turns modified. Table~\ref{tab:dialogue_validation_metrics} (Appendix~\ref{source_validation}) detailed analysis along with TER and CER. 

\subsection{BaatCheet\_Gold} \label{Benchmark_data}
\paragraph{Dialogue Sampling.}
For English source dialogues, we stratify a sample 90 dialogues each from \textit{DailyTalk}, \textit{MuTual}, and \textit{News-Eng*}. 
For Hindi, 90 dialogues each are sampled from \textit{Hinglish-Chat} and \textit{News-Hin*}. From each source, dialogues are partitioned into \textit{BaatCheet\_Human\_Train} and \textit{BaatCheet\_Test}. 
\begin{enumerate}[noitemsep,topsep=2pt,leftmargin=*]
    \item \textbf{\textit{BaatCheet\_Human\_Train}}: Used both as the reference set for Automatic metric-guided quality selection to identify the highest-quality translations for curating the synthetic training corpus, i.e., \textit{BaatCheet\_Synthetic} (detailed in section \ref{Train-Data}), and as gold training data for fine-tuning our selected translation models.
    \item \textbf{\textit{BaatCheet\_Test}}: Reserved exclusively as a held-out evaluation set for all translation experiments.
\end{enumerate}

These validated BaatCheet\_Gold source dialogues are human-translated by our inhouse translators. Translation guidelines, quality review processes, and detailed source validation metrics are described in Appendix~\ref{sec:baatcheet_pipeline_appendix}.

\subsection{Baselines Evaluation on BaatCheet\_Human\_Train} \label{mt_selection}

With our established human-translated \textit{BaatCheet\_Human\_Train}
reference set, we conduct a systematic evaluation of existing MT 
systems and open-source and proprietary LLMs on informal Indic dialogue translation 
across our target directions ($\text{Eng}\rightarrow\text{IL}$ and 
$\text{Hin}\rightarrow\text{IL}$). This evaluation serves a dual 
purpose: to test current models' capability on informal dialogue, and to identify which systems produce translations of sufficient quality to serve as candidate MT engines for large-scale synthetic data generation.

We evaluate nine systems on \textit{BaatCheet\_Human\_Train} using
reference-based COMET and reference-free COMET-QE across three categories: 
(i)~five open-source LLMs (\textit{Llama-3.1-8B-Instruct}\footnote{https://huggingface.co/meta-llama/Llama-3.1-8B-Instruct}~\cite{llama3_2_3b_instruct}, \textit{Llama-3.2-3B-Instruct}\footnote{https://huggingface.co/meta-llama/Llama-3.2-3B-Instruct}, \textit{Qwen3-4B-Instruct\footnote{https://huggingface.co/Qwen/Qwen3-4B-Instruct-2507}}~\cite{qwen3}, \textit{Gemma3-4B-IT}\footnote{https://huggingface.co/google/gemma-3-4b-it}~\cite{gemmateam2025gemma3technicalreport}, and \textit{Sarvam-Translate}~\footnote{https://huggingface.co/sarvamai/sarvam-translate}.); 
(ii)~three dedicated MT systems (\textit{Google Translate} (GMT)\footnote{\url{https://translate.google.co.in/?sl=auto\&tl=en\&op=translate}}, 
\textit{IndicTrans2}~\cite{gala2023indictrans2highqualityaccessiblemachine}, and \textit{BhashaVerse}\cite{mujadia2025bhashaversetranslationecosystem} \footnote{\url{https://ssmt.iiit.ac.in/translatev3}}); and 
(iii)~one proprietary LLM (\textit{GPT-4o-mini}~\cite{jiao2023chatgpt}).

\begin{figure*}[!t]
\centering
\includegraphics[width=\linewidth]{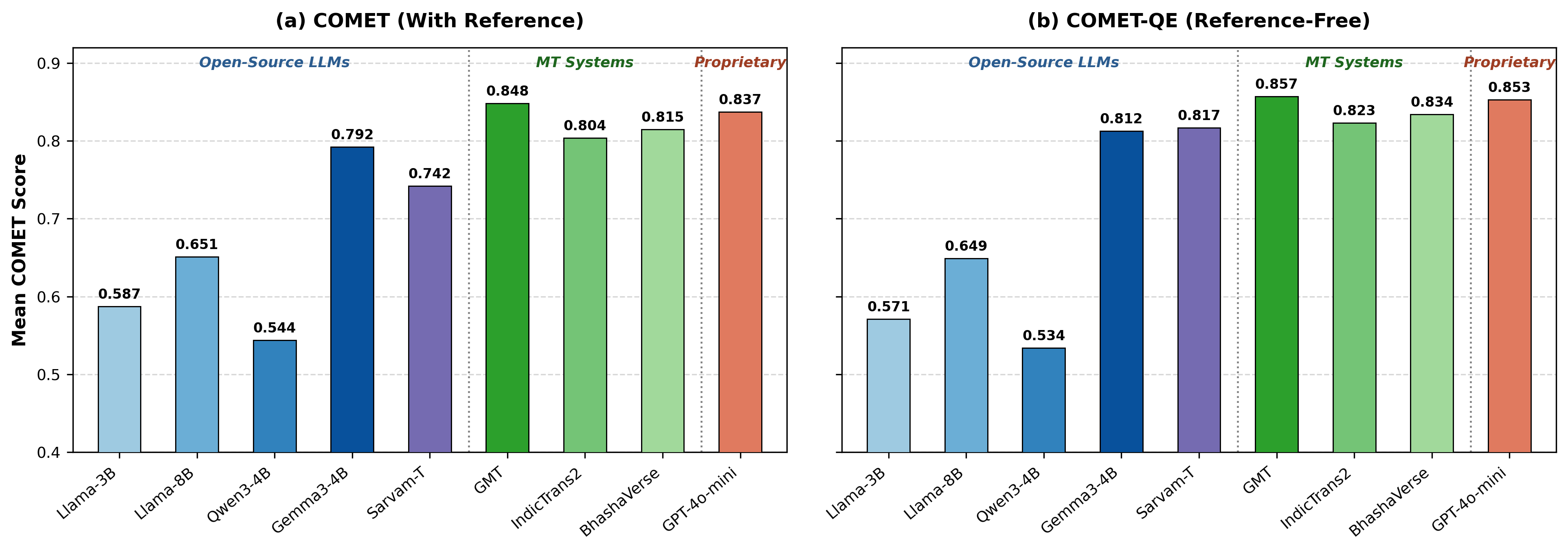}
\caption{Baseline performance comparison across nine candidate systems on \textit{BaatCheet\_Human\_Train} using (a)reference-based COMET and (b)reference-free COMET-QE. Dedicated MT systems (\textit{GMT}, \textit{IndicTrans2}, \textit{BhashaVerse}) and \textit{GPT-4o-mini} consistently outperform open-source LLMs across all language directions.}
\label{fig:baseline_mt_comet_comparison}
\end{figure*}

Figure~\ref{fig:baseline_mt_comet_comparison} reports scores across all five 
language directions. The three dedicated MT systems alongside \textit{GPT-4o-mini} 
consistently outperform among all models under both metrics (ref-based avg. 0.803--0.848; 
ref-free avg. 0.823--0.857). Among open-sourced LLMs, \textit{Gemma3-4B} and 
\textit{Sarvam-Translate} are the strongest, while \textit{Qwen3-4B} 
and \textit{Llama-3.2-3B} score lowest (avg. below 0.54 and 0.59 
respectively). These results reveal a clear quality gap between 
established MT systems and open-source LLMs for informal Indic 
dialogue translation, motivating our approach of using the four 
top-performing systems to generate large-scale synthetic training 
data, and subsequently fine-tuning the five open-source LLMs on this 
curated corpus to evaluate whether the gap can be closed.

\subsection{BaatCheet\_Synthetic Curation Pipeline} \label{Train-Data}
\paragraph{Synthetic Dialogue Sources.}
To enrich training data diversity and mitigate MT-bias on the Indic target side, we curate synthetic dialogues along two translation axes, as illustrated in Figure~\ref{fig:data_pipeline}.
\begin{enumerate}[noitemsep,topsep=2pt,leftmargin=*]
\item \textbf{$\text{Syn\_X2IL}$:} Comprises English source dialogues 
    \textit{DailyTalk}, \textit{MuTual}) and 
    \textit{News-Eng*} machine-translated into Hindi, Tamil, and Telugu; and Hindi source dialogues \textit{Hinglish-Chat}, \textit{News-Hin*} are machine-translated into 
    Tamil and Telugu.
 \item \textbf{$\text{Syn\_IL2X}$:} Comprises 
    Tamil dialogues (\textit{News-Tam*}, \textit{Story-Tam*}) and Telugu 
    dialogues (\textit{News-Tel*}, \textit{Story-Tel*}), machine-translated 
    into English and Hindi; and Hindi dialogues (\textit{Hinglish-Chat}, 
    \textit{News-Hin*}) machine-translated into English. All resulting pairs are 
    column-swapped to align with our target language pair direction X$\rightarrow$IL (English, Hindi to IL) training direction while fine-tuning our selected open-sourced LLMs, ensuring that the Indic target side in $\text{Syn\_IL2X}$ comprises dialogues generated natively in Indic style, preserving informal dialogue characteristics rather than just MT-biased output.
\end{enumerate}

\begin{table*}[th!]
\centering
\small
\setlength{\tabcolsep}{4.5pt}
\begin{tabular}{l rr rr c c@{\hskip 10pt}c c@{\hskip 10pt}c}
\toprule
\multirow{2}{*}{\textbf{BaatCheet Corpus}} & 
\multirow{2}{*}{\textbf{\#Dialogues}} & 
\multirow{2}{*}{\textbf{\#Utterances}} & 
\multicolumn{2}{c}{\textbf{\#Sentences}} & 
\multirow{2}{*}{\shortstack{\textbf{Avg Dlg}\\\textbf{Len}}} & 
\multicolumn{2}{c}{\textbf{Avg Utt Len}} & 
\multicolumn{2}{c}{\textbf{Avg Sent Len}} \\
\cmidrule(lr){4-5} \cmidrule(lr){7-8} \cmidrule(lr){9-10}
& & & \textbf{Src} & \textbf{Tgt} & & \textbf{Src} & \textbf{Tgt} & \textbf{Src} & \textbf{Tgt} \\
\midrule
\textit{BaatCheet\_Syn\_X2IL} & 42,148 & 301,278 & 533,102 & 533,494 & 8.8 & 14.8 & 11.8 & 8.7 & 6.9 \\
\textit{BaatCheet\_Syn\_IL2X} & 6,096  & 73,666  & 135,894 & 136,616 & 12.3 & 14.6 & 11.8 & 8.4 & 6.9 \\
\textit{BaatCheet\_Human\_Train}      & 510    & 8,405   & 16,455  & 16,077  & 17.3 & 17.6 & 14.1 & 9.1 & 7.5 \\
\textit{BaatCheet\_Test} & 658    & 8,904   & 16,813  & 16,518  & 13.7 & 16.9 & 14.1 & 8.9 & 7.5 \\
\midrule
\textbf{TOTAL}                 & \textbf{49,412} & \textbf{392,253} & \textbf{702,264} & \textbf{702,705} & \textbf{13.0} & \textbf{16.0} & \textbf{13.0} & \textbf{8.8} & \textbf{7.2} \\
\bottomrule
\end{tabular}
\caption{\small Overall statistics of the \textit{BaatCheet} corpus across 4 splits.}
\label{tab:summarized_stats}
\end{table*}

\paragraph{Selecting Candidate MT Systems.} \label{MT_canditates_Selection}
Based on our baseline evaluation on BaatCheet\_Human\_Train (Section~\ref{mt_selection}, Figure~\ref{fig:baseline_mt_comet_comparison}), the 
four top-performing systems, \textit{Google Translate} (GMT), 
\textit{GPT-4o-mini}, \textit{IndicTrans2}, and 
\textit{BhashaVerse} are selected as candidate MT outputs.

Each system independently translates all \textit{BaatCheet\_Human\_Train} 
dialogues, producing four parallel candidate translations per source 
utterance across both directions, forming 
\textit{Human\_Train\_X2IL} and \textit{Human\_Train\_IL2X} 
candidate sets. Language-pair disaggregated COMET scores of all four MT engines 
on these candidate sets are provided in 
Appendix~\ref{sec:appendix_candidate_comet}.

\paragraph{Automatic Metric Guided Quality Selection.}
We apply utterance-level reference-based (COMET) and reference-free (COMET-QE) scoring on 
\textit{Human\_Train\_X2IL} and \textit{Human\_Train\_IL2X} 
candidate sets to select the highest-quality translation per 
utterance from the four candidates, comparing two selection 
strategies:

\begin{enumerate}[noitemsep,topsep=1pt,leftmargin=*]
    \item \textbf{\textit{Utt-Avg} (Utterance-Average):} Selects the single MT system that achieves the highest average utterance score across an entire dialogue.
    \item \textbf{\textit{Utt-High} (Utterance-Highest):} Selects the highest-scoring candidate translation independently for every individual utterance within a dialogue.
\end{enumerate}

Analysis of MT candidate selection counts (Appendix~\ref{sec:appendix_mt_counts}) 
reveals that \textit{GMT} and \textit{GPT-4o-mini} are the most 
frequently selected engines under both strategies across all language directions.

\paragraph{Validation of Selection Strategies and Human Alignment.}
To determine which filtering strategy yields higher-quality translations on BaatCheet\_Human\_Train candidates, we compare Utt-Avg and Utt-High using DOC-COMET~\cite{vernikos2022embarrassingly} and COMTAIL~\cite{ahsan2025crosslingualoptimizedmetrictranslation}, plus a human binary preference study on 510 dialogues (255 per direction, X2IL and IL2X), where annotators chose between COMET-QE-curated Utt-Avg and Utt-High outputs. Both automatic metrics consistently favor Utt-High (Appendix~\ref{sec:appendix_filter_validation}), while humans preferred Utt-Avg in 58\% of dialogues (304/510) versus 42\% for Utt-High. This divergence is smaller than it appears: in over 90\% of dialogues, a single MT system contributes most utterances, so the two strategies select highly similar data. For example, in a 13-utterance dialogue (Appendix~\ref{sec:appendix_binary_preference}), Utt-High independently picks GMT for 8 turns, and GMT is also the top engine under Utt-Avg's dialogue-level average (0.8697). The two strategies converge on the same dominant engine even though candidate scoring is done turn-by-turn versus dialogue-wide. Disagreement concentrates in the remaining $\sim$10\%, where a dialogue-level average can be inflated by a few strong utterances that mask weaker ones (Appendix~\ref{sec:appendix_binary_preference}). Since our downstream goal is SFT data where turn-level accuracy directly shapes model output, we prioritize Utt-High's guaranteed per-utterance quality floor over Utt-Avg's holistic stylistic consistency. We therefore adopt \textit{Utt-High} for selecting the best-scoring translation at each turn via COMET-QE to construct \textit{BaatCheet\_Synthetic}.

\paragraph{BaatCheet Corpus Overview.}
Based on these findings, we curate the final \textit{BaatCheet\_Synthetic} dataset using reference-free COMET-QE under the \textit{Utt-High} strategy.
Table~\ref{tab:summarized_stats} summarizes \textit{BaatCheet} across its four splits, comprising 49,412 dialogues, 392,253 utterances, and $\sim$702k sentences. Large-scale synthetic data includes $\text{Syn\_X2IL}$ (42.1k dialogues) and $\text{Syn\_IL2X}$ (6.1k dialogues) with native Indic targets to reduce MT bias for our target direction $\text{X}\rightarrow\text{IL}$ fine-tuning of models. Human-translated dialogue splits contain 510 dialogues in $\text{Human\_Train}$ and 658 held-out dialogues in $\text{BaatCheet\_Test}$. Full breakdown appears in Table~\ref{tab:appendix_all_detailed_single_table}, Appendix~\ref{sec:appendix_stats}.
% Human dialogues exhibit greater turn depth (13.7--17.3 turns) and utterance length (16.9--17.6 words) than synthetic sets (8.8--12.3 turns), reflecting authentic conversational flow. 

\paragraph{Linguistics Properties.} Automatic language-ID tagging~\cite{madhani-etal-2023-bhasha} measures an average code-mixing rate of $\sim$20\% across language pairs, although this is likely an underestimate because it relies on Roman-script detection and misses the pervasive practice of transliterating English words into native Indic scripts. The corpus also preserves \textit{pro-drop} (null subjects/objects) and syntactic \textit{ellipsis}, especially Tamil and Telugu permit subjects and objects to be omitted when recoverable from the discourse context, as illustrated in Figure~\ref{fig:example_dia}. The translation challenge is not pro-drop itself, but correctly recovering omitted arguments when the target language requires explicit realization. Statistically evaluating this recovery is, however, not straightforward, as the Auto Pronoun Translation metric (APT~\cite{miculicich-werlen-popescu-belis-2017-validation}) is designed for pro-drop-scarce European languages and is unreliable in conversational Indic settings where omission is grammatical rather than exceptional\cite{hingrajiya-etal-2026-indicdisco}.

% The corpus also preserves \textit{pro-drop} (null subjects/objects) and syntactic \textit{ellipsis}, especially in the agglutinative Telugu and Tamil (Figure~\ref{fig:example_dia})\textemdash grammatical, not merely colloquial, features of these source languages, which is why we avoid pronoun-translation metrics such as APT~\cite{miculicich-werlen-popescu-belis-2017-validation} that were designed for pro-drop-scarce European languages. 
% Full analysis is in Appendix~\ref{sec:appendix_codemixing_prodrop}.

\section{BaatCheet Experimental Setup} \label{train}
\label{sec:methodology}
We select a structurally diverse suite of decoder-only instruction-tuned models: \textit{Llama-3.1-8B-Instruct}~\cite{llama3_2_3b_instruct}, \textit{Llama-3.2-3B-Instruct}, \textit{Qwen3-4B-Instruct}~\cite{qwen3}, \textit{Gemma3-4B-IT}~\cite{gemmateam2025gemma3technicalreport}, and \textit{Sarvam-Translate}~\footnote{sarvamai/sarvam-translate}. These models are chosen for their pre-training exposure to Indic languages, multilingual baselines, and optimized tokenizers for low-resource scripts. 

To establish strong baselines for comparing zero-shot and few-shot translation performance against supervised fine-tuning, we evaluate all models on \textit{BaatCheet\_Test} under two settings:
\begin{itemize}[noitemsep,topsep=2pt,leftmargin=*]
    \item \textbf{Zero-Shot Baseline:} Models translate dialogue prompts directly without in-context examples.
    \item \textbf{Few-Shot Baseline:} Implemented an adaptive retrieval strategy to dynamically select the most contextually relevant in-context examples. Specifically, we generate dense embeddings for all dialogues in the \textit{Human\_Train} set using the \texttt{all-MiniLM-L6-v2} model, which are indexed using FAISS. For each test dialogue, we retrieve the top-$k$ mathematically closest context examples. 
    
    % While this strategy improves translation quality as shown in Figure~\ref{fig:avg_metrics_LLMs}, we observe that several models are highly sensitive to prompt length, occasionally degrading translation performance as prompt context expands.
\end{itemize}

\begin{table*}[th!]
\centering
\small
\setlength{\tabcolsep}{2.5pt}
\renewcommand{\arraystretch}{1.1}
\resizebox{\textwidth}{!}{%
\begin{tabular}{l ccc cc ccc c}
\toprule
& \multicolumn{2}{c}{\textbf{Baselines}} 
& \multicolumn{2}{c}{\textbf{Synthetic-Only}} 
& \textbf{Human-Only} 
& \multicolumn{2}{c}{\textbf{Dual-Hybrid}} 
& \multicolumn{2}{c}{\textbf{Multi-Hybrid}} \\
\cmidrule(lr){2-3} \cmidrule(lr){4-5} \cmidrule(lr){6-6} \cmidrule(lr){7-8} \cmidrule(lr){9-10}
\textbf{Model} & \textbf{Zero-Shot} & \textbf{Few-Shot} & \textbf{Syn\_X2IL} & \textbf{Syn\_IL2X} & \textbf{HT} & \textbf{Syn\_X2IL+HT} & \textbf{Syn\_IL2X+HT} & \textbf{Syn\_Both+HT} & \textbf{Syn\_Both+3$\times$HT} \\
\midrule

\multicolumn{10}{c}{\textbf{(a) COMET (Reference-Based)}} \\
\midrule
Gemma3-4B    & 0.5024 & 0.6410 & 0.8076 & 0.7095 & \textbf{0.8324} & 0.8192 & 0.8223 & 0.7397 & 0.6087 \\
Llama-3.2-3B & 0.5317 & 0.5864 & \textbf{0.7802} & 0.7279 & 0.7139 & 0.7410 & 0.7076 & 0.7647 & 0.7271 \\
Llama-3.1-8B & 0.7113 & 0.6882 & 0.8023 & 0.7418 & 0.7871 & 0.7197 & 0.7210 & 0.7930 & \textbf{0.8461} \\
Qwen3-4B     & 0.6797 & 0.6601 & \textbf{0.8373} & 0.7372 & 0.7959 & 0.8063 & 0.7370 & 0.7695 & 0.8288 \\
Sarvam-T     & 0.6168 & 0.4859 & 0.8149 & 0.8057 & 0.8320 & \textbf{0.8573} & 0.8499 & 0.8471 & 0.8411 \\

\midrule
\multicolumn{10}{c}{\textbf{(b) DOC-COMET (Reference-Based)}} \\
\midrule
Gemma3-4B    & 0.4145 & 0.3755 & 0.7513 & 0.6342 & \textbf{0.7768} & 0.7673 & 0.7704 & 0.6666 & 0.5771 \\
Llama-3.2-3B & 0.5684 & 0.5152 & 0.6467 & 0.6113 & 0.6085 & 0.6418 & 0.6104 & \textbf{0.6622} & 0.6585 \\
Llama-3.1-8B & 0.6485 & 0.6067 & 0.6899 & 0.6418 & 0.7459 & 0.6265 & 0.6324 & 0.6799 & \textbf{0.7926} \\
Qwen3-4B     & 0.6008 & 0.5819 & \textbf{0.7846} & 0.6337 & 0.7408 & 0.7488 & 0.6716 & 0.7097 & 0.7843 \\
Sarvam-T     & 0.4954 & 0.3491 & 0.7604 & 0.7608 & 0.8023 & \textbf{0.8043} & \textbf{0.8043} & 0.7916 & 0.7955 \\

\midrule
\multicolumn{10}{c}{\textbf{(c) COMTAIL (Reference-Based)}} \\
\midrule
Gemma3-4B    & 0.2556 & 0.4791 & 0.7017 & 0.5655 & \textbf{0.7244} & 0.7007 & 0.6986 & 0.6011 & 0.4552 \\
Llama-3.2-3B & 0.3615 & 0.4169 & \textbf{0.6813} & 0.5835 & 0.5722 & 0.5999 & 0.5608 & 0.6596 & 0.6037 \\
Llama-3.1-8B & 0.5230 & 0.5489 & 0.7072 & 0.5918 & 0.6481 & 0.5711 & 0.5704 & 0.6945 & \textbf{0.7563} \\
Qwen3-4B     & 0.5648 & 0.5309 & \textbf{0.7506} & 0.5904 & 0.6719 & 0.6810 & 0.5956 & 0.6764 & 0.7372 \\
Sarvam-T     & 0.4723 & 0.3126 & 0.7202 & 0.6896 & 0.7328 & 0.7554 & 0.7443 & \textbf{0.7660} & 0.7582 \\

\bottomrule
\end{tabular}%
}
\caption{\small Performance comparison of five open-source LLMs across fine-tuning configurations on \textit{BaatCheet\_Test}, evaluated using (a) COMET DA, (b) DOC-COMET, and (c) COMTAIL DA. \texttt{Syn\_Both} denotes $\text{Syn\_[X2IL+IL2X]}$ and \texttt{HT} denotes $\text{Human\_Train}$. Bold values indicate the best-performing data configuration for each respective model.}
\label{tab:main_results_9configs}
\end{table*}

\paragraph{Fine-Tuning Setup.}
Models are loaded in $4$-bit NormalFloat (NF4) quantized format using \textit{bitsandbytes} \cite{dettmers2022llmint8} and trained using $\text{bfloat16}$ precision for numerical stability. The absolute context length is constrained to $4096$ tokens. To prevent truncation of long, multi-turn dialogue contexts, we implement a \textit{length-aware chunking} protocol: dialogues exceeding the maximum token budget are segmented into contiguous sub-dialogues, with each segment prefixed by a custom control token, \texttt{<splitted\_dialogue>}, to preserve contextual and structural continuity. This chunking mechanism is applied identically during both training and inference. The training prompt template is shown in Appendix section~\ref{train_prompt}.

\paragraph{Parameter-Efficient Fine-tuning.}
We employ Parameter-Efficient Fine-Tuning (PEFT) leveraging QLoRA combined with Rank-Stabilized LoRA (rsLoRA)~\cite{kalajdzievski2023rankstabilizationscalingfactor}. Specifically, rsLoRA enhances training stability across linguistically distinct target vocabularies by scaling the adapter updates with a factor proportional to $\alpha / \sqrt{r}$ (where $r$ denotes the LoRA rank and $\alpha$ is the scaling factor), rather than the traditional $\alpha / r$ scaling. LoRA adapter weights, particularly matrix A is Gaussian-initialized to encourage diverse feature representations and matrix B is initialized to zeros. The complete set of training hyperparameters is detailed Table~\ref{tab:hyperparameters} in the Appendix.

% We use QLoRA \cite{dettmers2023qlora} and Rank-Stabilized LoRA (rsLoRA) for fine-tuning. rsLoRA enhances stability by scaling updates with a factor proportional to $\alpha / \sqrt{r}$, ensuring stable gradient flow across linguistically diverse data. Adapter weights are Gaussian-initialized to encourage diverse feature representation. Full hyperparameters are provided in Appendix Table~\ref{tab:hyperparameters}.
% We adopt the QLoRA\cite{dettmers2023qlora} framework to perform parameter-efficient fine-tuning across all models. The detailed training hyperparameters are summarized in Table~[\ref{tab:hyperparameters} in the Appendix. In addition, we incorporate Rank-Stabilized LoRA (rsLoRA), which improves training stability when using higher LoRA ranks by scaling updates with a factor proportional to $\alpha / \sqrt{r}$. This helps maintain stable gradient flow across linguistically diverse training data. Adapter weights are initialized using a Gaussian distribution to encourage diverse feature representations during adaptation.

% To analyze the contribution of different data sources, we fine-tune the models on multiple dataset configurations derived from the \textit{BaatCheet} corpus. These include the synthetic datasets \textit{BaatCheet\_Syn\_X2IL} and \textit{BaatCheet\_Syn\_IL2X}, as well as the human-translated dataset \textit{BaatCheet\_Human\_Train}. 

\subsection{Training Data Configurations}

To systematically investigate the effect of dataset composition on informal dialogue translation quality, we define seven configurations spanning five categories:

\begin{enumerate}[noitemsep,topsep=2pt,leftmargin=*]
    \item \textbf{Baselines:} The zero-shot and few-shot baseline performance of each model.
    \item \textbf{Synthetic-Only:} Fine-tuning exclusively on each synthetic split (\texttt{Synt\_X2IL} vs.\ \texttt{Synt\_IL2X}) in isolation, to evaluate whether training on target-side machine translations (\texttt{Synt\_X2IL}) versus column-swapped native Indic target text (\texttt{Synt\_IL2X}) more effectively boosts fine-tuned translation performance.
    \item \textbf{Human-Only:} Fine-tuning strictly on gold standard translation set- \textit{Human\_Train}.
    \item \textbf{Dual-Hybrid:} Blending a single synthetic dialogue set with human gold standard - \textit{Syn\_X2IL +HT} and \textit{Syn\_IL2X +HT} to evaluate the complementary gain from combining machine-translated scale with human-translated dialogues.
    \item \textbf{Multi-Hybrid:} Blending all synthetic dialogue sets - \textit{Syn\_X2IL} and \textit{Syn\_IL2X} with human translation sets at two scaling ratios: \textit{Syn\_Both +HT} and \textit{Syn\_Both + 3$\times$HT} to assess whether increasing the proportion of human-translated data further improves translation quality at scale.
\end{enumerate}

% \paragraph{Evaluation Metrics.}
% All configurations are evaluated on \textit{BaatCheet\_Test} using 
% three complementary automatic metrics: \\
% (i)~\textbf{COMET}~\cite{rei-etal-2022-comet}, 
% measuring utterance-level semantic adequacy; \\
% (ii)~\textbf{DOC-COMET}, a context-aware extension that evaluates 
% cross-utterance coherence at the dialogue level; and \\
% (iii)~\textbf{COMTAIL}~\cite{ahsan2025crosslingualoptimizedmetrictranslation}, an Indic-specialized metric.

% \begin{figure*}[!]
%   \centering
%   \includegraphics[width=1\textwidth]{latex/sarvam_dialogue_bar_alignment.png}
%   \caption{Dialogue-Level 3-Factor SQM+DA and Utterance-Level Evaluation across language pairs. Grouped bars display $Z$-score rescaled ratings ($0\text{--}100$) for GPT, Gemini, and Human judges, demonstrating the consistent margin by which the pure \textit{Human\_Train} configuration outperforms the hybrid \textit{Syn\_X2IL +HT} setup across all dimensions.}
%   \label{fig:dialogue_bars}
% \end{figure*}

\section{Results and Discussion} \label{results}
We evaluate five fine-tuned LLMs across all dataset configurations on 
\textit{BaatCheet\_Test} using three automatic metrics 
(Table~\ref{tab:main_results_9configs}). Detailed results across all five LLMs on all configurations are provided in the Appendix Section~\ref{Results}.

\subsection{Metric Behaviour.}

\paragraph{COMET}~\cite{rei-etal-2022-comet} scores are consistently the highest across all 
configurations (ranging 0.49--0.86). Being a sentence-level
metric, COMET rewards local lexical and semantic adequacy per utterance 
independently, without modelling cross-turn discourse coherence. 
% This makes it relatively lenient in multi-turn dialogue settings where earlier context may be critical fo rcorrectness.

\paragraph{DOC-COMET}~\cite{vernikos2022embarrassingly} scores are systematically lower (0.35--0.79), as it 
incorporates up to three preceding utterances as additional context signal. 
While this enables coarse dialogue-level coherence scoring, its limited 
context window is insufficient to capture long-range dependencies in 
multi-turn conversations. Consequently, translations that are locally fluent but contextually incoherent are penalized, resulting in lower overall scores compared to sentence-level COMET.

\paragraph{COMTAIL}~\cite{ahsan2025crosslingualoptimizedmetrictranslation} produces the lowest scores overall (0.25–0.77). As an Indic-
specialized metric, COMTAIL may be more sensitive to Indic language-specific imperfections in lexical choice, morphology, and domain register. However, since COMTAIL is primarily fine-tuned on formal sentence-level parallel data, it may lack calibration for informal dialogue registers, resulting in conservatively lower scores.

\subsection{Training Data Configuration Observations}
\paragraph{Baselines.}
Few-shot performance is consistently lower than zero-shot for most models 
(e.g., \textit{Sarvam-T}: $0.6168 \rightarrow 0.4859$ COMET; 
\textit{Gemma}: $0.4145 \rightarrow 0.3755$ DOC-COMET). We attribute this 
to two compounding issues: (i)~retrieved in-context examples from 
\textit{BaatCheet\_Human\_Train} may not align well in topic or register with 
test dialogues, introducing distributional noise into the prompt; and (ii) 
multi-turn dialogue examples significantly expand input length, 
leading to context window saturation and truncation of earlier dialogue 
history, forcing the model to translate with incomplete context.
\paragraph{Synthetic-Only.}
Fine-tuning on $\text{Syn\_X2IL}$ consistently outperforms $\text{Syn\_IL2X}$ 
across all models and metrics. In $\text{Syn\_X2IL}$, the source side 
largely comprises naturally occurring dialogues (\textit{DailyTalk}, 
\textit{MuTual}, \textit{Hinglish-Chat}) with machine-translated Indic targets (Section~\ref{MT_canditates_Selection}). 
In contrast, $\text{Syn\_IL2X}$ is entirely synthetic on both sides: 
its source (English/Hindi) is machine-translated output, 
while its Indic targets are \textit{GPT-4o-mini}-generated dialogues 
which, despite being prompted for informal code-mixed style, lack 
the naturalistic register variation inherent in human-produced dialogues. 
Combined with its substantially smaller scale (6.1k vs.\ 42.1k dialogues), 
this fully synthetic construction limits the quality and diversity of 
the SFT signal in $\text{Syn\_IL2X}$.
\paragraph{Human-Only.}
Despite comprising only 510 dialogues, \textit{BaatCheet\_Human\_Train} alone 
achieves the highest COMET score for \textit{Gemma3-4B} (0.832) and 
competitive results for \textit{Sarvam} (0.832). Models with strong 
multilingual pre-training appear capable of efficiently leveraging small 
but high-quality human-translated corpora.
\paragraph{Dual-Hybrid.}
$\text{Syn\_X2IL}+\text{Human\_Train}$ yields peak performance for three of 
five models: \textit{Llama-3.2-3B} (0.765 COMET), \textit{Sarvam} 
(0.857), and \textit{Qwen} (0.806). $\text{Syn\_IL2X}+\text{Human\_Train}$, 
by contrast, consistently underperforms its counterpart across all 
models. Beyond the fully synthetic nature of $\text{Syn\_IL2X}$, its 
substantially smaller volume fails to add meaningful 
linguistic diversity to \textit{Human\_Train} during SFT.
\paragraph{Multi-Hybrid.}
$\text{Syn\_Both}+3\times\text{Human\_Train}$ delivers peak scores for 
\textit{Llama-3.1-8B} (COMET: 0.846, DOC-COMET: 0.793, COMTAIL: 0.756) 
and strong gains for \textit{Qwen3-4B} (COMET: 0.829 vs.\ 0.770 with 
$1\times\text{Human\_Train}$). However, $\text{Syn\_Both}+1\times\text{Human\_Train}$ 
noticeably underperforms noticeably \textit{hurts} these models relative to their 
$\text{Syn\_X2IL}+\text{Human\_Train}$ dual-hybrid performance in Gemma, which we 
attribute to a \textit{gold dilution effect}: the ${\sim}48$k combined 
synthetic volume overwhelms the 510-dialogue \textit{Human\_Train} set, 
exposing the model disproportionately to noisier synthetic signal. 
Tripling \textit{Human\_Train} ($3\times$) restores the quality-to-volume ratio, providing 
sufficient human-translated exposure for larger-capacity models to 
calibrate towards authentic informal dialogue quality.

% \paragraph{Effectiveness of Indic-Specialized Pre-training.}
% \textit{Sarvam-Translate} delivers the most consistently strong 
% performance across all data training configurations and metrics. Its DOC-COMET scores plateau across \textit{Human\_Train} and Dual-Hybrid scores, suggesting that its Indic-specialized pre-training helps it understand the task better. 

\begin{table*}[th!]
\centering
\scriptsize
\setlength{\tabcolsep}{2.5pt}
\renewcommand{\arraystretch}{1.15}
\resizebox{\textwidth}{!}{%
\begin{tabular}{ll c|ccc c|ccc c|ccc}
\toprule
& & \multicolumn{4}{c}{\textbf{GPT Judge}} 
  & \multicolumn{4}{c}{\textbf{Gemini Judge}} 
  & \multicolumn{4}{c}{\textbf{Human Judge}} \\
\cmidrule(lr){3-6} \cmidrule(lr){7-10} \cmidrule(lr){11-14}
& & \textbf{Utt-Level} & \multicolumn{3}{c}{\textbf{Dia-Level Factors}}
  & \textbf{Utt-Level} & \multicolumn{3}{c}{\textbf{Dia-Level Factors}}
  & \textbf{Utt-Level} & \multicolumn{3}{c}{\textbf{Dia-Level Factors}} \\
\cmidrule(lr){3-3} \cmidrule(lr){4-6} \cmidrule(lr){7-7} \cmidrule(lr){8-10} \cmidrule(lr){11-11} \cmidrule(lr){12-14}
\textbf{Model} & \textbf{Configuration} & \textbf{Acc} & \textbf{Nat} & \textbf{Inf} & \textbf{Coh} & \textbf{Acc} & \textbf{Nat} & \textbf{Inf} & \textbf{Coh} & \textbf{Acc} & \textbf{Nat} & \textbf{Inf} & \textbf{Coh} \\
\midrule

\multirow{2}{*}{Gemma}
& \texttt{Syn\_X2IL+HT} & 48.58 & 42.79 & 48.82 & 46.66 & 47.94 & 45.95 & 48.86 & 47.69 & 49.98 & 46.45 & 49.63 & 47.46 \\
& \texttt{Human\_Train} & 50.31 & 48.99 & 50.52 & 50.92 & 50.54 & 51.62 & 52.25 & 52.79 & 51.52 & 50.20 & 50.15 & 50.67 \\
\addlinespace[4pt]

\multirow{2}{*}{Llama-3B}
& \texttt{Syn\_X2IL+HT} & 43.14 & 39.55 & 46.76 & 37.47 & 41.31 & 37.49 & 41.81 & 37.13 & 43.22 & 43.11 & 49.23 & 42.31 \\
& \texttt{Syn\_X2IL} & 44.41 & 43.64 & 41.54 & 39.82 & 43.96 & 41.95 & 40.78 & 40.50 & 43.35 & 44.88 & 46.25 & 43.91 \\
\addlinespace[4pt]

\multirow{2}{*}{Llama-8B}
& \texttt{Syn\_Both+3$\times$HT} & 52.73 & 55.93 & 51.35 & 55.41 & 53.00 & 55.45 & 52.62 & 55.23 & 52.48 & 53.61 & 50.03 & 53.67 \\
& \texttt{Human\_Train} & 47.76 & 45.09 & 51.38 & 48.44 & 46.52 & 44.45 & 51.20 & 45.58 & 48.84 & 46.03 & 51.67 & 47.01 \\
\addlinespace[4pt]

\multirow{2}{*}{Qwen}
& \texttt{Syn\_Both+3$\times$HT} & 50.68 & 52.59 & 48.68 & 52.73 & 51.47 & 52.06 & 49.86 & 52.06 & 51.42 & 51.05 & 48.32 & 51.25 \\
& \texttt{Syn\_X2IL} & 51.86 & 52.61 & 47.80 & 52.66 & 51.73 & 52.46 & 49.93 & 52.60 & 51.92 & 51.30 & 48.50 & 51.56 \\
\addlinespace[4pt]

\multirow{2}{*}{Sarvam-T}
& \texttt{Syn\_X2IL+HT} & 53.99 & 58.28 & 55.64 & 57.30 & 54.86 & 58.18 & 56.38 & 57.55 & 54.76 & 55.38 & 52.90 & 55.32 \\
& \texttt{Human\_Train} & \textbf{54.76} & \textbf{60.55} & \textbf{57.62} & \textbf{58.66} & \textbf{56.18} & \textbf{60.59} & \textbf{57.32} & \textbf{59.20} & \textbf{55.53} & \textbf{58.04} & \textbf{53.92} & \textbf{57.18} \\
\bottomrule
\end{tabular}%
}
\caption{\small Qualitative SQM-DA evaluation scores ($z$-score normalized and scaled) disaggregated by \textbf{Utterance-Level Accuracy} (Utt-level) and \textbf{Dialogue-Level Factors} (Dia-level): Naturalness (\textbf{Nat}), Informality (\textbf{Inf}), and Cohesion \& Coherence (\textbf{Coh}). Evaluated across top-two data configurations per model by GPT-4o-mini, Gemini-Flash-Lite-preview, and Human Judges. \texttt{Syn\_Both} denotes $\text{Syn\_[X2IL+IL2X]}$. Bold text indicates the higher score per model setup.}
\label{tab:llm_human_sqm_da_combined_separated}
\end{table*}

% \paragraph{Impact of Fine-Tuned Data Configurations.}
% Figure~\ref{fig:avg_finetuned_configs} compares mean performance across fine-tuned setups (averaged over all models and metrics). $\text{Syn\_X2IL}$ achieves the highest overall score (0.7491), followed closely by human-only $\text{Human\_Train}$ (0.7323), outperforming all hybrid configurations. This demonstrates the effectiveness of utterance-highest (\textit{Utt-High}) COMET-based quality filtering. These results highlight the roles of synthetic and human data, where synthetic data provides linguistic coverage and human data anchors informal register style. 

\begin{figure}[h!]
\centering
\includegraphics[height=0.19\textheight]{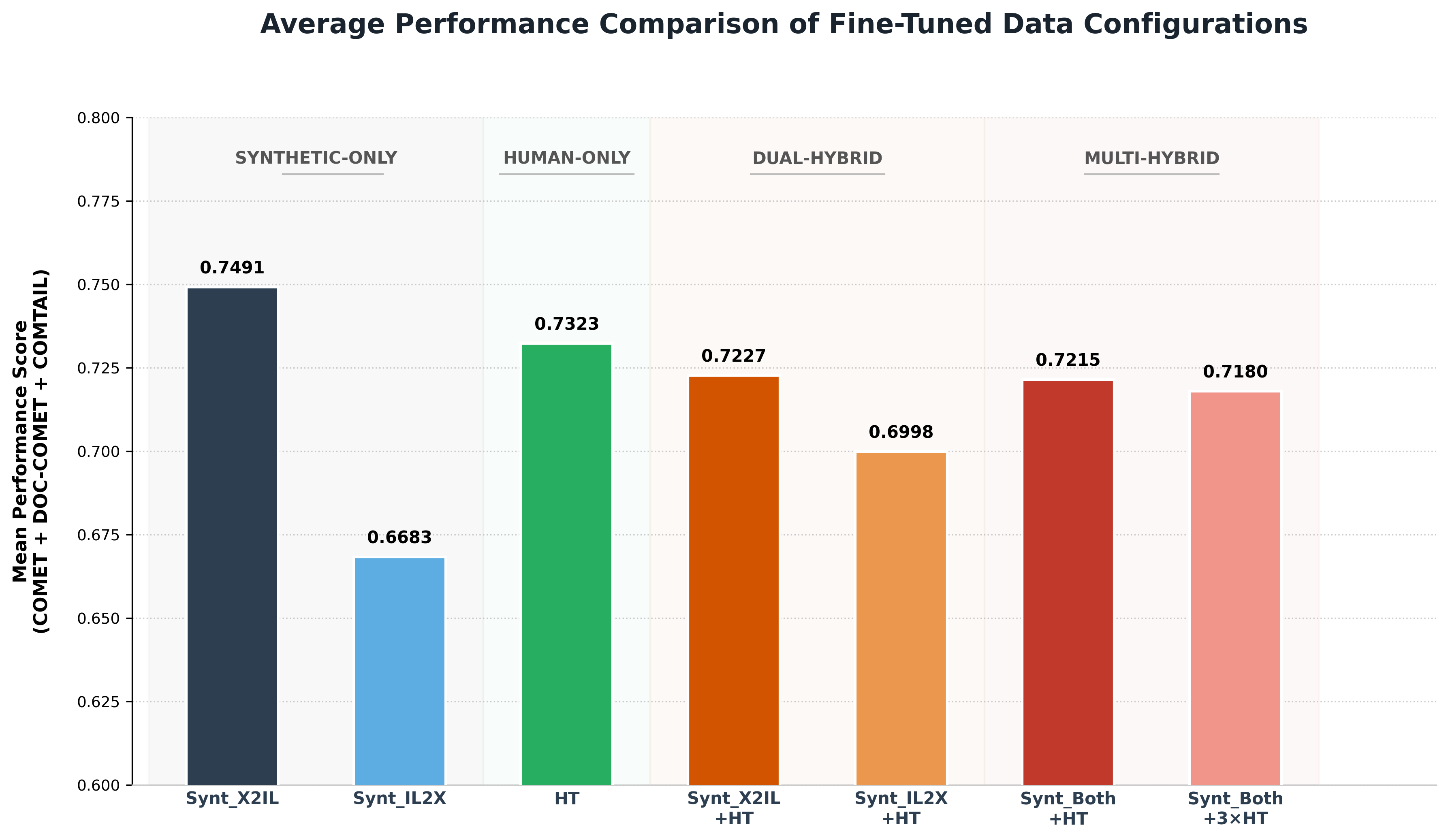}
\caption{\small Average translation performance (mean score across all five fine-tuned models and three evaluation metrics: COMET, DOC-COMET, and COMTAIL) for each fine-tuned training configuration.}
\label{fig:avg_finetuned_configs}
\end{figure}

\subsection{Comparative Analysis of Training Data Configurations}
To determine which data composition strategy most effectively boosts model 
performance, Figure~\ref{fig:avg_finetuned_configs} compares the mean 
scores across all fine-tuning configurations (averaged across all models 
and evaluation metrics). Among all configurations, \texttt{Synt\_X2IL} achieves the highest overall average score (\textbf{0.7491}), closely followed by the human-only baseline \texttt{Gold} (\textbf{0.7323}). Conversely, \texttt{Synt\_IL2X} in isolation yields the lowest overall performance (\textbf{0.6683}), and incorporating it into hybrid setups consistently degrades overall translation quality compared to \texttt{Synt\_X2IL} alone. Crucially, the responsiveness of each model to these data configurations 
is strongly governed by base architectural factors like model parameter 
capacity and pre-trained Indic linguistic knowledge.

\subsection{Limitations of Automatic Metrics}
While automatic metrics confirm substantial and consistent gains from 
our curated corpus, they do not fully capture pragmatic dimensions 
critical to informal dialogue translation, including honorific pronoun usage, pro-drop nature, and code-mixed naturalness. 
% To validate these findings beyond automatic scoring, we additionally conduct LLM-based and expert human evaluation, discussed in Section~\ref {llm_da}.

% While automatic metrics correlate in preserving global dialogue coherence, they struggle with register nuances like naturalness, informality and coherence.
% To bridge these limitations and select configurations for qualitative human evaluation, we average sentence COMET and context DOC-COMET scores (Figure~\ref{fig:zeroshot_final_scores}). Under this joint ranking, \textit{Sarvam-Translate} fine-tuned on Gold standard (\textit{BaatCheet\_Human\_Train}, $0.8172$) and the hybrid mixture (\textit{Syn\_X2IL + Gold]}, $0.8308$) emerged as the top two performing setups. We select these configurations for close-up LLM-as-a-judge: GPT-5-mini and Gemini-3.1-flash-lite-preview and human expert evaluations using a Scalar Quality Metric (SQM)-guided Direct Assessment (DA) protocol on a continuous 0 to 100 scale. We sample 30 dialogues per English-source pair and 20 dialogues per Hindi-source pair, totaling 260 dialogues (180 English-source and 80 Hindi-source). The detailed scoring guidelines and evaluators recurting protocals are detailed in Appendix~\ref{appendix_normalization}.

\section{LLM-as-Judge and Human Assessment} \label{llm_da}
We conduct a qualitative evaluation using both human judges and LLM judges under a Scalar Quality Metric (SQM)-guided Direct Assessment (DA) protocol. 
\paragraph{Experimental Setup and Sampling.}
For each of the five fine-tuned models, we select its top-two performing training configurations based on DOC-COMET scores, yielding 10 unique model-configuration setups. From each setup, we sample test dialogues across all language directions: 30 dialogues for each English-source pair ($\text{Eng}\rightarrow\text{Hin/Tam/Tel}$) and 20 dialogues for the Hindi-source pair ($\text{Hin}\rightarrow\text{Tel}$). In total, 1,100 dialogues (900 English-source and 200 Hindi-source) are independently evaluated by expert native-speaking linguists and two SOTA LLM judges: \textit{GPT-5-mini} and \textit{Gemini-3.1-flash-lite-preview}.
\paragraph{Scoring Granularity.}
Evaluations are conducted at two distinct levels:
\textbf{Dialogue Level:} The judge assesses the full context on three dimensions: \textit{Naturalness}, \textit{Informality}, and \textit{Cohesion \& Coherence}.
\textbf{Utterance Level:} Each sentence pair is independently scored to capture fine-grained translation accuracy. 
The detailed prompt templates and guidelines are given in section~\ref{appendix_normalization} in the Appendix.

\subsection{Qualitative Assessment and Analysis} \label{sec:qualitative_analysis}
\paragraph{Cross-Judge Agreement.}
Table~\ref{tab:llm_human_sqm_da_combined_separated} reveals strong directional consistency across all three evaluators, GPT, Gemini, 
and Human, with judges largely agreeing on both the 
relative ranking of models and the preferred training configuration per 
model. This concordance validates the reliability of LLM judges as 
proxies for human assessment in the informal Indic dialogue translation 
setting.

\paragraph{Model-Specific Qualitative Trends.}
Disaggregating scores by model architecture reveals distinct qualitative behaviors:
\begin{itemize}[noitemsep,topsep=2pt,leftmargin=*]
    \item For \textit{Gemma3-4B} and \textit{Sarvam-T}, the \texttt{Human\_Train} configuration consistently yields higher scores than \texttt{Syn\_X2IL + HT} across all judges and all four evaluation dimensions: utterance accuracy, naturalness, informality, and cohesion. This is consistent with 
their automatic metric results (Section~\ref{results}).
    \item \textit{Llama-3.2-3B} records the lowest qualitative scores across all 
judges and dimensions, mirroring its modest gains under automatic metrics. 
This consistent under-performance across both evaluation paradigms 
confirms that its 3B parameter capacity remains the primary bottleneck, 
limiting its ability to capture discourse structure regardless 
of the training data configuration.
    
    \item For \textit{Llama-3.1-8B} and \textit{Qwen3-4B}, both evaluated 
configurations produce closely competitive qualitative scores across all 
judges, with neither configuration showing a decisive advantage over the 
other in dialogue-level factors. Notably, for these two models, 
Gemini and Human judgements exhibit stronger mutual 
agreement compared to GPT, suggesting that Gemini's evaluation 
tendencies better align with human preferences for dialogue-level 
naturalness and coherence assessment in this setting.
\end{itemize}

\paragraph{Overall SQM-guided DA Analysis.}
Across all models and judges, \textit{Sarvam-T} under \texttt{Human\_Train} 
achieves the highest qualitative scores on all three dialogue-level 
factors, particularly Naturalness (GPT: 60.55; Gemini: 60.59; Human: 
58.04) and Cohesion (GPT: 58.66; Gemini: 59.20; Human: 57.18). 
This is fully consistent with its consistently strong performance in automatic evaluation (Section~\ref{results}).

\paragraph{Granular Rubric Distribution for Sarvam-T.}
At the dialogue level, 85/130 instances were rated as \textit{Grammatical and somewhat natural}, while 64/130 were classified as \textit{Slightly Informal}. Regarding discourse flow, 66/130 dialogues were deemed \textit{Somewhat coherent and cohesive}, with 52/130 achieving \textit{Strong Coherence and Cohesion}. At the utterance level, over 90\% of dialogues achieved high accuracy and naturalness (scores $\ge 61$). A detailed, granular analysis per language pair, along with a sample dialogue error analysis, is provided in Table~\ref{tab:sarvam_sqm_distribution_combined} and Figure~\ref{fig:sarvam_dialogue_sample} in Appendix~\ref{Human_DA}.

\begin{table}[t!]
\centering
\small
\setlength{\tabcolsep}{3.5pt}
\renewcommand{\arraystretch}{1.1}
\begin{tabular}{l cc cc cc}
\toprule
& \multicolumn{2}{c}{\textbf{GPT vs H}} 
& \multicolumn{2}{c}{\textbf{Ge vs Hu}} 
& \multicolumn{2}{c}{\textbf{GPT vs Ge}} \\
\cmidrule(lr){2-3} \cmidrule(lr){4-5} \cmidrule(lr){6-7}
\textbf{Factor} & $r$ & $\tau$ & $r$ & $\tau$ & $r$ & $\tau$ \\
\midrule
Nat           & 0.552 & 0.370 & 0.625 & 0.428 & 0.738 & 0.540 \\
Inf           & 0.343 & 0.232 & 0.332 & 0.209 & 0.391 & 0.266 \\
Coh & 0.561 & 0.364 & 0.647 & 0.442 & 0.743 & 0.532 \\
Utt       & \textbf{0.702} & \textbf{0.498} & \textbf{0.697} & \textbf{0.525} & \textbf{0.781} & \textbf{0.597} \\
\bottomrule
\end{tabular}
\caption{\small Average inter-judge correlation ($r$ = Pearson, $\tau$ = Kendall's $\tau$) across all five models. Bold values indicate the highest agreement factor per judge pair(GPT, Gemini-\textbf{Ge} and Human-\textbf{H}.\textbf{Utterance-Level Accuracy} (Utt-level) and \textbf{Dialogue-Level Factors} (Dia-level): Naturalness (\textbf{Nat}), Informality (\textbf{Inf}), and Cohesion \& Coherence (\textbf{Coh}). \textbf{Ge}.}
\label{tab:inter_judge_correlation_avg}
\end{table}

\subsection{Inter-Judge Correlation Analysis}
Table~\ref{tab:inter_judge_correlation_avg} reports average inter-judge 
correlations across all five models. Utterance-level accuracy achieves 
the strongest agreement across all judge pairs, confirming that 
fine-grained translation accuracy is consistently assessable by both 
human and LLM evaluators. However, when evaluating 
holistic dialogue-level factors, correlation drops noticeably across all 
evaluators, while Informality records the lowest 
correlations. This reveals a fundamental evaluation gap: while judges reliably agree on local-context utterance-level factors, evaluating on dialogue-level factors introduces subjectivity and the cultural-sensitive nature of Judges. This shows that holistic dialogue quality is far more challenging to standardize than utterance-level accuracy. Full per-model breakdowns are provided in Appendix~\ref{sec:appendix_correlation}.

\section{Conclusion} \label{concl}
We introduce \textit{\textbf{BaatCheet}}, a multi-turn parallel dialogue corpus for informal Indic translation across five language directions. Our evaluation reveals that while dedicated MT engines and \textit{GPT-4o-mini} lead zero-shot baselines, fine-tuning open-source LLMs—particularly Indic-specialized models like \textit{Sarvam-T} significantly bridges this gap. We demonstrate that synthetic data curated from natural source dialogues (\texttt{Synt\_X2IL}) provides a far superior training signal compared to synthetic Indic sources (\texttt{Synt\_IL2X}). Through our evaluations, we show that LLMs can be employed as judges for evaluating dialogue translations as proxies. Future extensions of this study will expand the size of BaatCheet\_Human dialogues also to include Bengali, Gujarati, Kannada, and Marathi, maintaining the current source language pairs. Additionally, we aim to develop domain-specific datasets to further refine model performance for specialized downstream applications like healthcare.

\section*{Limitations}
Synthetic target generation and MT translation may introduce subtle structural artifacts, lacking the spontaneous code-mixing and deep cultural nuances of organic human conversations. Human-translated gold data, while highly effective for fine-tuning, is limited in volume due to annotation costs, making dataset expansion a key priority. Systemic scoring differences between LLM judges (e.g., Gemini vs.\ GPT) 
require calibration against human baselines, and budget constraints limited 
human Direct Assessment to a sampled subset of 1,100 dialogues.

\section*{Acknowledgments}
This work is supported by HIMANGY (HIndustani Machini ANuvaad TechnoloGY), a consortium-based initiative under BHASHINI, funded by the
Ministry of Electronics and Information Technology (MeitY), Government of India. We gratefully acknowledge their support in funding in creating the BaatCheet Corpus.
\bibliography{custom}

\appendix

\section{Appendix}
\label{sec:appendix}

\section{Formal vs Informal Translation}

\begin{figure*}[ht!]
    \centering
    \includegraphics[height=0.3\textheight]{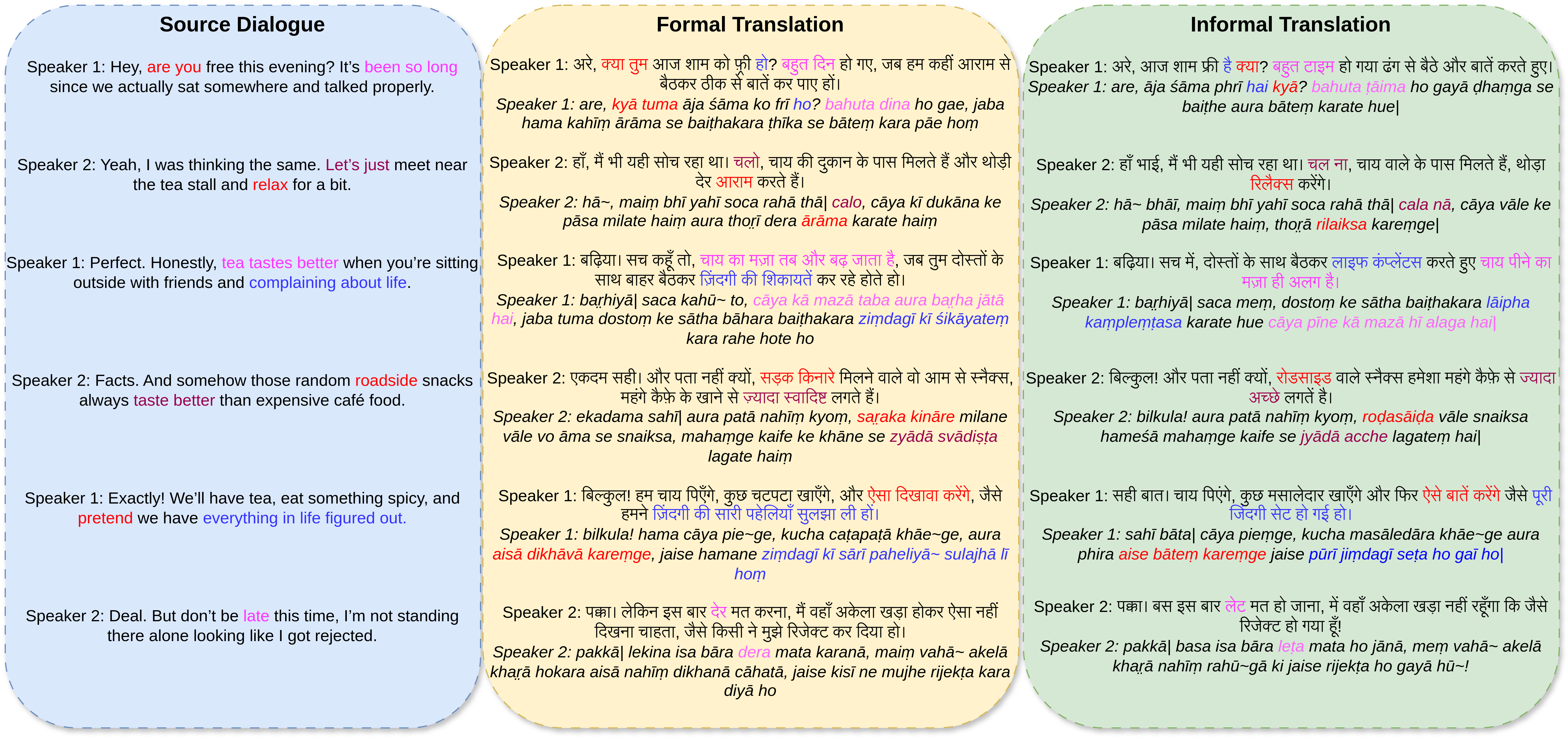}
    \caption{An example of Formal and Informal Translation from English to Hindi along with translitration in IAST format. All the differences are highlited with colors.}
    \label{fig:formal_informal_examples}
\end{figure*}

Formal language relies on rigid grammatical structures and standardized vocabulary, whereas informal "BaatCheet" prioritizes social flow and spontaneity through pro-drop (omitted subjects) and code-mixing. For instance, in given example Figure~\ref{fig:formal_informal_examples}, the formal translation uses explicit pronouns and literary metaphors like paheliyan sulajhana (solving riddles), which feel robotic in a casual setting. Conversely, the informal translation naturalizes the dialogue by adding social markers like "Bhai" and utilizing common urban slang like "life set" instead of a literal translation for "figured out." This shift demonstrates that informal translation is a process of pragmatic transcreation, capturing the "vibe" and cultural nuances that formal, sentence-level MT systems often miss.

\section{BaatCheet Corpora Building}
 \label{corpora_building}
 Table~\ref{tab:data_sources_description} presents the overview of dialogue data sources used for constructing BaatCheet Corpora across English, Hindi, Tamil, and Telugu.

\begin{table*}[th!]
\centering
\small
\renewcommand{\arraystretch}{1.25}
\begin{tabularx}{\textwidth}{p{0.28\textwidth} X}
\toprule
\textbf{Data Source -- Src. Lang} & \textbf{Description} \\
\midrule
\textbf{Hinglish-Chat} -- Hin &
A synthetically generated Hinglish conversational dataset~\cite{Hinglish-Chat-21M} featuring everyday-life dialogues.
The dataset is modified by LLM prompting to \textit{``Remove unnatural noun and pronoun usage from this dialogue, but it still has to make sense and write it in Hindi (Devanagari) script.''} \\

\textbf{MuTual} (Multi-Turn Dialogue Reasoning) -- Eng &
A Chinese high school English listening comprehension dataset~\cite{mutual}, adapted, preprocessed, and manually annotated to create a multi-turn dialogue reasoning corpus aimed at evaluating dialogue models' contextual understanding and reasoning capabilities. \\

\textbf{DailyTalk} -- Eng &
A conversational speech dataset for TTS systems developed by sampling and modifying dialogues from the open-domain DailyDialog corpus~\cite{li-etal-2017-dailydialog}. Dialogues from DailyTalk~\cite{lee2022dailytalk} are used here, annotated with speaker identifiers to capture natural conversational flow. \\

\textbf{News Articles} -- Eng, Hin, Tam, Tel &
News articles~\cite{mujadia2025bhashaversetranslationecosystem} from diverse domains are used to generate informal dialogues using LLM prompting:
\textit{``Generate 10 informal dialogues between two friends on this topic. Keep them short, natural, conversational and direct, and conclude meaningfully.''} \\

\textbf{Short Stories} -- Tam, Tel &
Short stories from online sources such as \texttt{tamil\_stories}\footnote{\url{https://huggingface.co/datasets/aitamilnadu/tamil_stories}} for Tamil and the \texttt{Chandamama Kathalu Dataset}\footnote{\url{https://code.swecha.org/telugu-ai/chandamama-kathalu-dataset}} for Telugu are used. Each story is prompted as:
\textit{``You are given a [language] story text. Generate one informal dialogue between two [language] friends discussing the story in detail in a natural, code-mixed (with English) tone in [language]. The dialogue should be conversational and informal like spoken language. Reach an opinion about the story.''} \\
\bottomrule
\end{tabularx}
\caption{\small Overview of dialogue data sources used for constructing BaatCheet Corpora across English, Hindi, Tamil, and Telugu.}
\label{tab:data_sources_description}
\end{table*}

%%%%%%%%%%%%%%%%%%%%%%%%%%%%%%%%%%%%%%%%%%%%%%%%%%%%%
\subsection{Guidelines for Generated Dialogue Validation} \label{source_validation}
Language experts are provided with the following standardized instructions to validate LLM-generated dialogues:
\paragraph{Validation Guidelines}
\begin{enumerate}
\item Grammatical correctness and semantic coherence.
\item Contextual pronoun consistency across dialogue turns.
\item Gender agreement and awkward phrasing remediation.
\item Converting grammatically sound yet robotic or overly formal expressions into natural, colloquial style.
\item Refining code-mixed phrasing to ensure conversational naturalness.
\end{enumerate}

\paragraph{Metric Calculation}
\paragraph{Dialogue Edit Rate (\%):} The percentage of dialogues where at least one Utterance was modified by human: $$\text{Dialogue Edit Rate} = \frac{N_{\text{edited dialogues}}}{N_{\text{total dialogues}}} \times 100$$

\paragraph{Utterance Edit Rate (\%):} The proportion of individual dialogue Utterances altered during validation: $$\text{Utterance Edit Rate} = \frac{T_{\text{edited}}}{T_{\text{total}}} \times 100$$

\paragraph{Token Edit Rate (TER \%):} Corpus-level word-level Levenshtein edit distance (insertions, deletions, substitutions) normalized by total source word count: $$\text{TER} = \frac{\sum_{i=1}^{T} \text{Levenshtein}_{\text{word}}(\text{Source}_i, \text{Target}i)}{\sum{i=1}^{T} \text{WordCount}(\text{Source}_i)} \times 100$$

\paragraph{Character Edit Rate (CER \%):} Corpus-level character-level Levenshtein edit distance normalized by total source character length: $$\text{CER} = \frac{\sum_{i=1}^{T} \text{Levenshtein}_{\text{char}}(\text{Source}_i, \text{Target}i)}{\sum{i=1}^{T} \text{CharCount}(\text{Source}_i)} \times 100$$

\begin{table*}[th!]
\centering
\small
\setlength{\tabcolsep}{6pt} % Reduce column padding
\resizebox{\textwidth}{!}{%
\begin{tabular}{l c c c c c c c c}
\toprule
\textbf{Datasource} & \textbf{\begin{tabular}[c]{@{}c@{}}Total\\ Dialogues\end{tabular}} & \textbf{\begin{tabular}[c]{@{}c@{}}Edited\\ Dialogues\end{tabular}} & \textbf{\begin{tabular}[c]{@{}c@{}}Dialogue Edit\\ Rate (\%)↓\end{tabular}} & \textbf{\begin{tabular}[c]{@{}c@{}}Total\\ Utts\end{tabular}} & \textbf{\begin{tabular}[c]{@{}c@{}}Edited\\ Utts\end{tabular}} & \textbf{\begin{tabular}[c]{@{}c@{}}Utt Edit\\ Rate(\%)↓\end{tabular}} & \textbf{\begin{tabular}[c]{@{}c@{}}TER↓\\ (\%)\end{tabular}} & \textbf{\begin{tabular}[c]{@{}c@{}}CER↓\\ (\%)\end{tabular}} \\
\midrule
News-Eng$^\dagger$ & 50 & 4 & 8.00 & 516 & 4 & 0.78 & 0.31 & 0.25 \\
Hinglish$^\dagger$ & 49 & 49 & 100.00 & 967 & 638 & 65.98 & 21.99 & 19.60 \\
News-Hin$^\dagger$ & 49 & 34 & 69.39 & 492 & 139 & 28.25 & 3.96 & 2.77 \\
News-Tam & 48 & 27 & 56.25 & 344 & 61 & 17.73 & 4.81 & 2.94 \\
News-Tel & 48 & 38 & 79.17 & 361 & 167 & 46.26 & 17.87 & 11.91 \\
Stories-Tam & 50 & 48 & 96.00 & 777 & 330 & 42.47 & 9.68 & 6.21 \\
Stories-Tel & 50 & 47 & 94.00 & 1,030 & 332 & 32.23 & 6.88 & 4.43 \\
\midrule
\textbf{TOTAL} & \textbf{344} & \textbf{247} & \textbf{71.80} & \textbf{4,487} & \textbf{1,671} & \textbf{37.24} & \textbf{10.47} & \textbf{7.63} \\
\bottomrule
\end{tabular}%
}
\caption{Edit metrics of Human Validation across all data sources. $^\dagger$Indicates BaatCheet\_Gold source dialogues; remaining splits are sampled from trainset splits. We report Dialogue Edit Rate, Utterance(Utt) Edit Rate, Token Edit Rate (TER), and Character Edit Rate (CER). Lower is better.}
\label{tab:dialogue_validation_metrics}
\end{table*}

\subsection{Detailed BaatCheet Corpus Statistics}
\label{sec:appendix_stats}
The entire \textit{BaatCheet} corpus aggregates to 49,412 dialogues, 392,253 utterances, and over 702,000 parallel sentences as shown in Table~\ref{tab:appendix_all_detailed_single_table}.
\begin{itemize}[noitemsep,topsep=2pt,leftmargin=*]
    \item \textbf{Synthetic Volume ($\text{Syn\_X2IL}$ \& $\text{Syn\_IL2X}$):} The synthetic component provides the core scale necessary for LLM fine-tuning. $\text{BaatCheet\_Syn\_X2IL}$ accounts for 42,148 dialogues (301,278 utterances), representing standard $\text{X}\rightarrow\text{IL}$ translations. $\text{BaatCheet\_Syn\_IL2X}$ contributes 6,096 dialogues (73,666 utterances), which are column-swapped to provide natively generated Indic targets and eliminate machine translation bias on the target side.
    \item \textbf{Human Gold Standards ($\text{Human\_Train}$ \& $\text{BaatCheet\_Test}$):} The professionally translated human sets consist of 510 dialogues (8,405 utterances) in $\text{BaatCheet\_Human\_Train}$ and 658 dialogues (8,904 utterances) in $\text{BaatCheet\_Test}$. $\text{Human\_Train}$ acts as both a high-fidelity fine-tuning set and the quality-filtering reference for synthetic curation, while $\text{BaatCheet\_Test}$ is strictly held out for evaluation.
\end{itemize}

\paragraph{Diversity in Linguistic Properties}
Comparing Source vs. Target utterance lengths (in words) and sentence lengths reveals fundamental linguistic properties across language families: \\
\textbf{Agglutination (Tamil \& Telugu).}
In $\text{X}\rightarrow\text{IL}$ directions ($\text{Eng}\rightarrow\text{Tam}$ and $\text{Eng}\rightarrow\text{Tel}$), English source utterances average 14.97--17.37 words, whereas Tamil and Telugu target utterances average only 11.18--12.76 words. Similarly, in $\text{IL}\rightarrow\text{X}$ directions ($\text{Tam}\rightarrow\text{Eng}$ and $\text{Tel}\rightarrow\text{Eng}$), Tamil and Telugu source utterances (9.59--12.12 words) expand significantly when translated into English targets (11.86--16.50 words). 
\begin{quote}
\textit{Linguistic Cause:} Tamil and Telugu are highly agglutinative Dravidian languages. Case markers, postpositions, tense inflections, and pronominal suffixes are morphologically bound into single word forms, resulting in systematically lower word counts compared to analytic English.
\end{quote}

\paragraph{Domain Diversity}
By combining daily speech (\textit{DailyTalk}), multi-turn reasoning (\textit{MuTual}), code-mixed chat (\textit{Hinglish-Chat}), generated news conversations (\textit{News}), and fictional narratives (\textit{Stories}), \textit{BaatCheet} covers a broad spectrum of registers from task-oriented information exchange to informal chitchat.

\begin{table*}[h]
\centering
\scriptsize
\setlength{\tabcolsep}{3.5pt}
\renewcommand{\arraystretch}{0.98}
\resizebox{\textwidth}{!}{%
\begin{tabular}{ll rr c cc cc}
\toprule
\textbf{Direction} & \textbf{Data Source} & \textbf{\#Dialogues} & \textbf{\#Utterances} & \textbf{\#Sentences (Src / Tgt)} & \textbf{Avg Dlg Len} & \textbf{Avg Utt Len (Src / Tgt)} & \textbf{Avg Sent Len (Src / Tgt)} \\
\midrule

% =============================================================================
% (a) BaatCheet_Synt_X2IL
% =============================================================================
\multicolumn{8}{c}{\textbf{\small (a) BaatCheet\_Syn\_X2IL} ($\text{X}\rightarrow\text{IL}$ Direction: English / Hindi $\longrightarrow$ Indic Language)} \\
\midrule
\multirow{4}{*}{Eng$\rightarrow$Hin} 
& DailyTalk   & 2,446  & 22,152 & 32,582 / 32,312 & 9.06  & 11.28 / 11.41 & 7.81 / 7.94 \\
& MuTual      & 8,346  & 38,332 & 68,560 / 68,594 & 4.59  & 16.35 / 17.56 & 9.29 / 9.93 \\
& News        & 826    & 7,834  & 12,690 / 12,663 & 9.48  & 11.86 / 13.39 & 7.34 / 8.30 \\
& \textbf{Total} & \textbf{11,618} & \textbf{68,318} & \textbf{113,832 / 113,569} & \textbf{5.88} & \textbf{14.97 / 15.97} & \textbf{8.84 / 9.39} \\
\addlinespace[2pt]
\multirow{4}{*}{Eng$\rightarrow$Tam} 
& DailyTalk   & 2,446  & 22,152 & 32,582 / 32,585 & 9.06  & 11.28 / 8.52  & 7.81 / 5.90 \\
& MuTual      & 8,346  & 38,332 & 68,560 / 68,819 & 4.59  & 16.35 / 12.31 & 9.29 / 6.97 \\
& News        & 826    & 7,834  & 12,690 / 12,676 & 9.48  & 11.86 / 8.98  & 7.34 / 5.56 \\
& \textbf{Total} & \textbf{11,618} & \textbf{68,318} & \textbf{113,832 / 114,080} & \textbf{5.88} & \textbf{14.97 / 11.28} & \textbf{8.84 / 6.64} \\
\addlinespace[2pt]
\multirow{4}{*}{Eng$\rightarrow$Tel} 
& DailyTalk   & 2,446  & 22,152 & 32,582 / 32,668 & 9.06  & 11.28 / 8.52  & 7.81 / 5.87 \\
& MuTual      & 8,346  & 38,332 & 68,560 / 68,698 & 4.59  & 16.35 / 12.18 & 9.29 / 6.89 \\
& News        & 826    & 7,834  & 12,690 / 12,706 & 9.48  & 11.86 / 8.95  & 7.34 / 5.53 \\
& \textbf{Total} & \textbf{11,618} & \textbf{68,318} & \textbf{113,832 / 114,072} & \textbf{5.88} & \textbf{14.97 / 11.18} & \textbf{8.84 / 6.58} \\
\addlinespace[2pt]
\multirow{3}{*}{Hin$\rightarrow$Tam} 
& HinglishGit & 1,558  & 33,349 & 75,054 / 75,101 & 21.41 & 17.12 / 12.46 & 7.47 / 5.43 \\
& News        & 2,089  & 14,813 & 20,749 / 20,823 & 7.09  & 12.78 / 8.71  & 9.27 / 6.29 \\
& \textbf{Total} & \textbf{3,647} & \textbf{48,162} & \textbf{95,803 / 95,924} & \textbf{13.21} & \textbf{14.63 / 10.31} & \textbf{8.50 / 5.92} \\
\addlinespace[2pt]
\multirow{3}{*}{Hin$\rightarrow$Tel} 
& HinglishGit & 1,558  & 33,349 & 75,054 / 75,057 & 21.41 & 17.12 / 12.65 & 7.47 / 5.51 \\
& News        & 2,089  & 14,813 & 20,749 / 20,792 & 7.09  & 12.78 / 8.74  & 9.27 / 6.32 \\
& \textbf{Total} & \textbf{3,647} & \textbf{48,162} & \textbf{95,803 / 95,849} & \textbf{13.21} & \textbf{14.63 / 10.41} & \textbf{8.50 / 5.97} \\

\midrule
% =============================================================================
% (b) BaatCheet_Synt_IL2X (Flipped Src / Tgt for IL -> X)
% =============================================================================
\multicolumn{8}{c}{\textbf{\small (b) BaatCheet\_Syn\_IL2X} ($\text{IL}\rightarrow\text{X}$ Direction: Indic Language [Src] $\longrightarrow$ English / Hindi [Tgt])} \\
\midrule
\multirow{3}{*}{Hin$\rightarrow$Eng} 
& HinglishGit & 671    & 14,332 & 32,177 / 32,121 & 21.36 & 17.16 / 15.49 & 7.50 / 6.78 \\
& News        & 957    & 7,055  & 9,719 / 9,747   & 7.37  & 12.68 / 11.36 & 9.31 / 8.29 \\
& \textbf{Total} & \textbf{1,628} & \textbf{21,387} & \textbf{41,896 / 41,868} & \textbf{13.14} & \textbf{14.52 / 13.06} & \textbf{8.56 / 7.67} \\
\addlinespace[2pt]
\multirow{3}{*}{Tam$\rightarrow$Eng} 
& News        & 775    & 5,568  & 8,665 / 8,697   & 7.18  & 10.18 / 13.64 & 6.61 / 8.15 \\
& Stories     & 591    & 9,025  & 17,904 / 17,888 & 15.27 & 12.09 / 16.50 & 6.14 / 8.10 \\
& \textbf{Total} & \textbf{1,366} & \textbf{14,593} & \textbf{26,569 / 26,585} & \textbf{11.23} & \textbf{11.13 / 15.07} & \textbf{6.37 / 8.13} \\
\addlinespace[2pt]
\multirow{3}{*}{Tel$\rightarrow$Eng} 
& News        & 335    & 2,705  & 3,650 / 3,720   & 8.07  & 9.59 / 11.86  & 7.11 / 8.59 \\
& Stories     & 533    & 8,842  & 17,141 / 16,520 & 16.59 & 12.12 / 15.35 & 6.22 / 8.15 \\
& \textbf{Total} & \textbf{868} & \textbf{11,547} & \textbf{20,791 / 20,240} & \textbf{13.30} & \textbf{11.14 / 14.00} & \textbf{6.56 / 8.32} \\
\addlinespace[2pt]
\multirow{3}{*}{Tam$\rightarrow$Hin} 
& News        & 775    & 5,568  & 8,665 / 8,697   & 7.18  & 10.18 / 14.01 & 6.61 / 9.05 \\
& Stories     & 591    & 9,025  & 17,904 / 17,888 & 15.27 & 12.09 / 17.75 & 6.14 / 9.01 \\
& \textbf{Total} & \textbf{1,366} & \textbf{14,593} & \textbf{26,569 / 26,585} & \textbf{11.23} & \textbf{11.14 / 15.63} & \textbf{6.37 / 9.03} \\
\addlinespace[2pt]
\multirow{3}{*}{Tel$\rightarrow$Hin} 
& News        & 335    & 2,705  & 3,650 / 3,569   & 8.07  & 9.59 / 12.38  & 7.11 / 9.33 \\
& Stories     & 533    & 8,841  & 17,141 / 17,047 & 16.59 & 12.12 / 17.12 & 6.22 / 8.85 \\
& \textbf{Total} & \textbf{868} & \textbf{11,546} & \textbf{20,791 / 20,616} & \textbf{13.30} & \textbf{11.14 / 15.29} & \textbf{6.56 / 9.03} \\
\midrule

% =============================================================================
% (c) BaatCheet_Gold
% =============================================================================
\multicolumn{8}{c}{\textbf{\small (c) BaatCheet\_Human\_Train}} \\
\midrule
\multirow{4}{*}{Eng$\rightarrow$Hin} 
& DailyTalk   & 39     & 723    & 1,302 / 1,259   & 18.54 & 15.87 / 16.46 & 8.98 / 9.58 \\
& MuTual      & 40     & 447    & 901 / 895       & 11.18 & 21.76 / 22.85 & 9.56 / 9.98 \\
& News        & 39     & 476    & 888 / 748       & 12.21 & 14.38 / 18.07 & 7.67 / 11.68 \\
& \textbf{Total} & \textbf{118} & \textbf{1,646} & \textbf{3,091 / 2,902} & \textbf{13.95} & \textbf{17.37 / 19.16} & \textbf{8.74 / 10.41} \\
\addlinespace[2pt]
\multirow{4}{*}{Eng$\rightarrow$Tam} 
& DailyTalk   & 39     & 723    & 1,302 / 1,339   & 18.54 & 15.87 / 11.50 & 8.98 / 6.30 \\
& MuTual      & 40     & 447    & 901 / 907       & 11.18 & 21.76 / 15.44 & 9.56 / 6.84 \\
& News        & 39     & 476    & 888 / 850       & 12.21 & 14.38 / 11.27 & 7.67 / 6.33 \\
& \textbf{Total} & \textbf{118} & \textbf{1,646} & \textbf{3,091 / 3,096} & \textbf{13.95} & \textbf{17.37 / 12.76} & \textbf{8.74 / 6.49} \\
\addlinespace[2pt]
\multirow{4}{*}{Eng$\rightarrow$Tel} 
& DailyTalk   & 39     & 723    & 1,302 / 1,329   & 18.54 & 15.87 / 12.12 & 8.98 / 6.71 \\
& MuTual      & 40     & 447    & 901 / 920       & 11.18 & 21.76 / 14.92 & 9.56 / 6.53 \\
& News        & 39     & 476    & 888 / 802       & 12.21 & 14.38 / 11.17 & 7.67 / 6.78 \\
& \textbf{Total} & \textbf{118} & \textbf{1,646} & \textbf{3,091 / 3,051} & \textbf{13.95} & \textbf{17.37 / 12.75} & \textbf{8.74 / 6.67} \\
\addlinespace[2pt]
\multirow{3}{*}{Hin$\rightarrow$Tam} 
& HinglishGit & 39     & 1,369  & 3,023 / 3,015   & 35.10 & 20.82 / 14.90 & 9.32 / 6.68 \\
& News        & 39     & 379    & 568 / 570       & 9.72  & 15.07 / 9.82  & 9.73 / 6.33 \\
& \textbf{Total} & \textbf{78} & \textbf{1,748} & \textbf{3,591 / 3,585} & \textbf{22.41} & \textbf{17.94 / 12.36} & \textbf{9.53 / 6.50} \\
\addlinespace[2pt]
\multirow{3}{*}{Hin$\rightarrow$Tel} 
& HinglishGit & 39     & 1,369  & 3,024 / 2,891   & 35.10 & 20.82 / 15.66 & 9.32 / 7.38 \\
& News        & 39     & 379    & 568 / 552       & 9.72  & 15.07 / 10.98 & 9.73 / 7.46 \\
& \textbf{Total} & \textbf{78} & \textbf{1,748} & \textbf{3,592 / 3,443} & \textbf{22.41} & \textbf{17.94 / 13.32} & \textbf{9.53 / 7.42} \\

\midrule
% =============================================================================
% (d) BaatCheet_Benchmark
% =============================================================================
\multicolumn{8}{c}{\textbf{\small (d) BaatCheet\_Test}} \\
\midrule
\multirow{4}{*}{Eng$\rightarrow$Hin} 
& DailyTalk   & 51     & 862    & 1,430 / 1,458   & 16.90 & 15.53 / 16.23 & 9.58 / 9.85 \\
& MuTual      & 50     & 584    & 1,220 / 1,184   & 11.68 & 23.05 / 25.56 & 10.47 / 11.50 \\
& News        & 51     & 526    & 997 / 973       & 10.31 & 14.73 / 17.46 & 7.77 / 9.42 \\
& \textbf{Total} & \textbf{152} & \textbf{1,972} & \textbf{3,647 / 3,615} & \textbf{12.97} & \textbf{17.73 / 19.71} & \textbf{9.26 / 10.25} \\
\addlinespace[2pt]
\multirow{4}{*}{Eng$\rightarrow$Tam} 
& DailyTalk   & 51     & 862    & 1,430 / 1,465   & 16.90 & 15.53 / 11.42 & 9.58 / 6.86 \\
& MuTual      & 50     & 584    & 1,220 / 1,232   & 11.68 & 22.93 / 17.88 & 10.40 / 7.96 \\
& News        & 51     & 526    & 997 / 987       & 10.31 & 14.73 / 11.58 & 7.77 / 6.16 \\
& \textbf{Total} & \textbf{152} & \textbf{1,972} & \textbf{3,647 / 3,684} & \textbf{12.97} & \textbf{17.69 / 13.60} & \textbf{9.24 / 6.99} \\
\addlinespace[2pt]
\multirow{4}{*}{Eng$\rightarrow$Tel} 
& DailyTalk   & 51     & 862    & 1,430 / 1,344   & 16.90 & 15.53 / 11.60 & 9.58 / 7.66 \\
& MuTual      & 50     & 584    & 1,220 / 1,234   & 11.68 & 23.05 / 17.55 & 10.47 / 7.70 \\
& News        & 51     & 526    & 997 / 968       & 10.31 & 14.73 / 11.93 & 7.77 / 6.49 \\
& \textbf{Total} & \textbf{152} & \textbf{1,972} & \textbf{3,647 / 3,546} & \textbf{12.97} & \textbf{17.73 / 13.67} & \textbf{9.26 / 7.28} \\
\addlinespace[2pt]
\multirow{3}{*}{Hin$\rightarrow$Tam} 
& HinglishGit & 50     & 981    & 2,106 / 2,101   & 19.62 & 16.11 / 11.93 & 7.33 / 5.43 \\
& News        & 51     & 513    & 830 / 846       & 10.06 & 15.28 / 11.83 & 9.51 / 7.21 \\
& \textbf{Total} & \textbf{101} & \textbf{1,494} & \textbf{2,936 / 2,947} & \textbf{14.79} & \textbf{15.69 / 11.88} & \textbf{8.43 / 6.33} \\
\addlinespace[2pt]
\multirow{3}{*}{Hin$\rightarrow$Tel} 
& HinglishGit & 50     & 981    & 2,106 / 1,929   & 19.62 & 16.11 / 12.09 & 7.33 / 6.06 \\
& News        & 51     & 513    & 830 / 797       & 10.06 & 15.28 / 11.05 & 9.51 / 7.22 \\
& \textbf{Total} & \textbf{101} & \textbf{1,494} & \textbf{2,936 / 2,726} & \textbf{14.79} & \textbf{15.69 / 11.57} & \textbf{8.43 / 6.65} \\

\bottomrule
\end{tabular}%
}
\caption{\small Comprehensive statistics for all four \textit{BaatCheet} splits: (a)~\textit{BaatCheet\_Syn\_X2IL}, (b)~\textit{BaatCheet\_Syn\_IL2X}, (c)~\textit{BaatCheet\_Human\_Train}, and (d)~\textit{BaatCheet\_Test}.}
\label{tab:appendix_all_detailed_single_table}
\end{table*}

\subsection{Linguistic Properties: Code-Mixing \& Pronoun Ellipsis of BaatCheet Corpus} \label{sec:appendix_codemixing_prodrop}
\paragraph{Code-Mixing.}
To evaluate the prevalence of code-mixing in our corpus, we processed the datasets using the \textit{Indic ILD}~\cite{madhani-etal-2023-bhasha} language detection tool. The tool identifies an average code-mixing rate of approximately 20\% across our language pairs. However, we note that this baseline is a significant underestimate due to a crucial limitation of current automatic tools: Indic ILD relies primarily on the presence of the Roman script (English characters) to identify code-mixed tokens. In contrast, real-world colloquial Indic speech features extensive code-mixing using English words transliterated directly into native Indic scripts. Consequently, although the corpus exhibits a high density of conversational code-mixing, its true frequency cannot be precisely quantified by script-dependent automatic tools.
\paragraph{Pronoun-Drop and Ellipsis.}
Prior work in discourse-aware MT has proposed the Auto Pronoun Translation (APT~\cite{miculicich-werlen-popescu-belis-2017-validation}, which is specific for European Languages) metric to evaluate pronoun translation quality. We argue that adapting APT as a primary evaluation metric is highly unreliable in conversational Indic settings. Tamil and Telugu, like other pro-drop languages, permit subjects and
objects to be omitted when they are recoverable from grammatical or
discourse context. In dialogue translation, the challenge is therefore not
the presence of pro-drop itself, but correctly recovering omitted information
when translating into a language that requires or prefers its explicit
realization as shown in the figure~\ref{fig:example_dia}. 

\subsection{BaatCheet\_Gold Translation Protocal}
\label{sec:baatcheet_pipeline_appendix}
Prior to human translation, synthetically generated source dialogues of \textit{BaatCheet\_Gold} underwent expert editing to ensure naturalness, formality, and coherence, and the edit metrics are shown in Table~\ref {tab:dialogue_validation_metrics} in section~\ref{source_validation}.

\paragraph{Translation Protocol.}
Validated source dialogues were assigned to professional human translators with native-level proficiency in the target language and at least three years of translation experience. Per language pair, 2--4 translators independently produced target-language translations following strict informal dialogue guidelines.
\begin{enumerate}
 \item~manual translation only (no MT post-editing); 
 \item~use of colloquial vocabulary; avoidance of formal or Sanskritized 
registers; 
 \item~contextual adaptation of pronouns and honorifics based on speaker relationships; and 
 \item~preservation of natural turn-taking flow and discourse coherence. 
\end{enumerate}

Completed translations underwent an expert review by senior linguists (distinct from the translators) in the first pass; expert reviewers flagged translations violating any guideline; flagged segments were returned to the translator for revision for targeted revision. 
This iterative process ensures that \textit{BaatCheet\_Gold} constitutes a high-quality gold standard for informal dialogue MT evaluation.

\section{Baseline Candidate System Evaluation: Disaggregated Results}
\label{sec:appendix_baseline_mt}

Table~\ref{tab:appendix_candidate_systems_full} presents the complete language-pair disaggregated evaluation of all nine candidate systems on \textit{BaatCheet\_Human\_Train}, comparing reference-based COMET and reference-free COMET-QE scores across all five language directions ($\text{Eng}\rightarrow\text{Hin}$, $\text{Eng}\rightarrow\text{Tam}$, $\text{Eng}\rightarrow\text{Tel}$, $\text{Hin}\rightarrow\text{Tam}$, and $\text{Hin}\rightarrow\text{Tel}$).

\paragraph{LLM Translation Prompt Template.}
For zero-shot baseline evaluation of open-source LLMs (\textit{Llama-3B}, \textit{Llama-8B}, \textit{Qwen3-4B}, \textit{Gemma3-4B}, \textit{Sarvam-Translate}) and \textit{GPT-4o-mini}, we use a simple, direct instruction prompt without in-context examples:

\begin{tcolorbox}[title=Baseline Evaluation Prompt Template]  \label{train_prompt}
Translate the following from \{source\_language\} to \{target\_language\}: \{source\_text\}
\end{tcolorbox}
This simple template ensures that baseline evaluations measure pure zero-shot translation capabilities without prompt-engineering artifacts.

\paragraph{High-Resource Alignment ($\text{Eng}\rightarrow\text{Hin}$).}
$\text{Eng}\rightarrow\text{Hin}$ exhibits the highest scores across all open-source LLMs under both reference-based COMET (0.699--0.804) and reference-free COMET-QE (0.751--0.846). This superior baseline stems from the abundance of English-Hindi parallel training data in open-source LLM pre-training mixtures, enabling stronger zero-shot cross-lingual transfer even for informal dialogue registers.

\paragraph{Dravidian Language Challenge ($\text{Eng/Hin}\rightarrow\text{Tam/Tel}$).}
For Dravidian target languages (Tamil and Telugu), open-source LLMs experience a sharp performance drop, particularly for Hindi-source directions ($\text{Hin}\rightarrow\text{Tam}$ and $\text{Hin}\rightarrow\text{Tel}$). For instance, \textit{Llama-3.2-3B} drops from 0.728 on $\text{Eng}\rightarrow\text{Hin}$ to 0.505 on $\text{Hin}\rightarrow\text{Tam}$, while \textit{Qwen3-4B} drops from 0.699 to 0.508. This drop is attributed to morphological complexity (agglutination) and lower representation of Dravidian script in general open-source LLM pre-training.

\paragraph{Robustness of Dedicated MT Systems and GPT-4o-mini.}
In contrast to open-source LLMs, dedicated MT engines (\textit{Google Translate}, \textit{IndicTrans2}, \textit{BhashaVerse}) and \textit{GPT-4o-mini} demonstrate remarkable stability across all language pairs, maintaining high COMET scores ($>0.80$) even for Dravidian targets. \textit{Google Translate} (GMT) achieves peak overall average performance under both reference-based COMET (0.848) and COMET-QE (0.857), followed closely by \textit{GPT-4o-mini} (0.837 and 0.853). This cross-lingual stability justifies using these top four engines as candidate translation generators in our synthetic curation pipeline.

\begin{table*}[th!]
\centering
\small
\setlength{\tabcolsep}{3pt}
\renewcommand{\arraystretch}{1.1}
\resizebox{\textwidth}{!}{%
\begin{tabular}{l ccccc cccc}
\toprule
& \multicolumn{5}{c}{\textbf{Open-Source LLMs}} 
& \multicolumn{3}{c}{\textbf{Dedicated MT Systems}} 
& \textbf{Proprietary LLM} \\
\cmidrule(lr){2-6} \cmidrule(lr){7-9} \cmidrule(lr){10-10}
\textbf{Lang Pair} & \textbf{Llama-3B} & \textbf{Llama-8B} & \textbf{Qwen3-4B} & \textbf{Gemma3-4B} & \textbf{Sarvam-T} & \textbf{GMT} & \textbf{IndicTrans2} & \textbf{BhashaVerse} & \textbf{GPT-4o-mini} \\
\midrule

\multicolumn{10}{c}{\textbf{(a) Reference-Based COMET}} \\
\midrule
Eng$\rightarrow$Hin & 0.7282 & 0.7546 & 0.6989 & 0.7881 & 0.8038 & 0.7957 & 0.7524 & 0.7543 & 0.7857 \\
Eng$\rightarrow$Tam & 0.5745 & 0.7141 & 0.4945 & 0.8109 & 0.7725 & \textbf{0.8623} & 0.8079 & 0.8100 & 0.8393 \\
Eng$\rightarrow$Tel & 0.5964 & 0.6257 & 0.5023 & 0.7644 & 0.8056 & \textbf{0.8391} & 0.8296 & 0.8210 & 0.8255 \\
Hin$\rightarrow$Tam & 0.5054 & 0.5675 & 0.5079 & 0.7998 & 0.6541 & \textbf{0.8791} & 0.8010 & 0.8554 & 0.8686 \\
Hin$\rightarrow$Tel & 0.5316 & 0.5939 & 0.5164 & 0.7983 & 0.6748 & 0.8654 & 0.8273 & 0.8327 & \textbf{0.8659} \\
\midrule
\textbf{TOTAL AVG}  & 0.5872 & 0.6512 & 0.5440 & 0.7923 & 0.7422 & \textbf{0.8483} & \textbf{0.8036} & \textbf{0.8147} & \textbf{0.8370} \\

\midrule
\multicolumn{10}{c}{\textbf{(b) Reference-Free COMET-QE}} \\
\midrule
Eng$\rightarrow$Hin & 0.7745 & 0.8137 & 0.7507 & 0.8383 & 0.8463 & \textbf{0.8613} & 0.8430 & 0.8427 & 0.8498 \\
Eng$\rightarrow$Tam & 0.5446 & 0.7118 & 0.4892 & 0.8324 & 0.7741 & \textbf{0.8704} & 0.8462 & 0.8514 & 0.8597 \\
Eng$\rightarrow$Tel & 0.5589 & 0.6081 & 0.4914 & 0.8087 & 0.8184 & \textbf{0.8589} & 0.8517 & 0.8462 & 0.8519 \\
Hin$\rightarrow$Tam & 0.4637 & 0.5417 & 0.4485 & 0.7771 & 0.8160 & 0.8475 & 0.7710 & 0.8143 & \textbf{0.8532} \\
Hin$\rightarrow$Tel & 0.5130 & 0.5687 & 0.4893 & 0.8052 & 0.8284 & 0.8477 & 0.8041 & 0.8147 & \textbf{0.8500} \\
\midrule
\textbf{TOTAL AVG}  & 0.5710 & 0.6488 & 0.5338 & 0.8124 & 0.8166 & \textbf{0.8571} & \textbf{0.8232} & \textbf{0.8339} & \textbf{0.8529} \\

\bottomrule
\end{tabular}%
}
\caption{\small Language-pair disaggregated baseline evaluation across nine candidate systems on \textit{BaatCheet\_Human\_Train}, evaluated using (a) reference-based COMET and (b) reference-free COMET-QE. Bold text indicates the highest score per language pair.}
\label{tab:appendix_candidate_systems_full}
\end{table*}

\subsection{Candidate MT Evaluation: Disaggregated Results}
\label{sec:appendix_candidate_comet}

Table~\ref{tab:appendix_candidate_mt_combined} presents the complete language-pair and source-disaggregated evaluation of the four candidate MT engines (\textit{Google Translate}, \textit{GPT-4o-mini}, \textit{IndicTrans2}, and \textit{BhashaVerse}) on \textit{Human\_Train\_X2IL} and \textit{Human\_Train\_IL2X} candidate sets.

\paragraph{Engine Performance Trends.}
\textit{Google Translate} (GMT) and \textit{GPT-4o-mini} consistently emerge as the top-two performing engines across both translation axes. On $\text{X2IL}$ directions ($\text{X}\rightarrow\text{IL}$), GMT achieves peak average scores on English-source pairs ($\text{Eng}\rightarrow\text{Tam}$ COMET-QE: 0.8704), while \textit{GPT-4o-mini} leads on Hindi-source Indic pairs ($\text{Hin}\rightarrow\text{Tel}$ COMET: 0.8659). On $\text{IL2X}$ reverse directions ($\text{IL}\rightarrow\text{X}$), \textit{GPT-4o-mini} displays exceptional translation strength into English targets ($\text{Tel}\rightarrow\text{Eng}$ COMET: 0.8711).

\paragraph{Source-Level Insights.}
Conversational corpora (\textit{DailyTalk}, \textit{MuTual}) and news dialogues (\textit{News-gen}) maintain high translation quality across engines ($>0.80$ COMET). Conversely, code-mixed dialogues (\textit{Hinglish-Chat}) present greater cross-lingual difficulty when translating into Dravidian targets ($\text{Tam}\rightarrow\text{Hin}$ IndicTrans2 COMET: 0.6911), reflecting the morphological and syntactic challenges of informal code-mixed translation. These disaggregated results validate our selection of GMT and \textit{GPT-4o-mini} as the primary translation generators for synthetic corpus curation.

\begin{table*}[th!]
\centering
\scriptsize
\setlength{\tabcolsep}{3.5pt}
\renewcommand{\arraystretch}{1.06}
\resizebox{\textwidth}{!}{%
\begin{tabular}{ll cccc cccc}
\toprule
& & \multicolumn{4}{c}{\textbf{Reference-Based COMET}} 
  & \multicolumn{4}{c}{\textbf{Reference-Free COMET-QE}} \\
\cmidrule(lr){3-6} \cmidrule(lr){7-10}
\textbf{Lang Pair} & \textbf{Data Source} & \textbf{GMT} & \textbf{GPT} & \textbf{IndicTrans2} & \textbf{BhashaVerse} & \textbf{GMT} & \textbf{GPT} & \textbf{IndicTrans2} & \textbf{BhashaVerse} \\
\midrule

% =============================================================================
% (a) BaatCheet_Gold_X2IL
% =============================================================================
\multicolumn{10}{c}{\textbf{(a) 
Human\_Train\_X2IL ($\text{X}\rightarrow\text{IL}$ Direction)}} \\
\midrule

\multirow{4}{*}{Eng$\rightarrow$Hin} 
& DailyTalk    & \textbf{0.7628} & 0.7506 & 0.7350 & 0.7257 & \textbf{0.8587} & 0.8474 & 0.8417 & 0.8297 \\
& MuTual       & 0.8059 & \textbf{0.8064} & 0.7526 & 0.7724 & \textbf{0.8597} & 0.8492 & 0.8329 & 0.8440 \\
& News-gen     & \textbf{0.8185} & 0.8000 & 0.7696 & 0.7649 & \textbf{0.8655} & 0.8527 & 0.8542 & 0.8544 \\
& AVG          & \textbf{0.7957} & 0.7857 & 0.7524 & 0.7543 & \textbf{0.8613} & 0.8498 & 0.8430 & 0.8427 \\
\addlinespace[2pt]

\multirow{4}{*}{Eng$\rightarrow$Tam} 
& DailyTalk    & \textbf{0.8295} & 0.7900 & 0.8055 & 0.8010 & \textbf{0.8693} & 0.8502 & 0.8603 & 0.8563 \\
& MuTual       & \textbf{0.8607} & 0.8535 & 0.7890 & 0.8101 & \textbf{0.8690} & 0.8605 & 0.8258 & 0.8473 \\
& News-gen     & \textbf{0.8969} & 0.8743 & 0.8294 & 0.8189 & \textbf{0.8729} & 0.8684 & 0.8525 & 0.8505 \\
& AVG          & \textbf{0.8623} & 0.8393 & 0.8079 & 0.8100 & \textbf{0.8704} & 0.8597 & 0.8462 & 0.8514 \\
\addlinespace[2pt]

\multirow{4}{*}{Eng$\rightarrow$Tel} 
& DailyTalk    & 0.8390 & 0.7951 & \textbf{0.8409} & 0.8293 & \textbf{0.8607} & 0.8490 & 0.8544 & 0.8487 \\
& MuTual       & 0.8267 & \textbf{0.8342} & 0.8160 & 0.8065 & \textbf{0.8549} & 0.8489 & 0.8457 & 0.8405 \\
& News-gen     & \textbf{0.8517} & 0.8474 & 0.8319 & 0.8272 & \textbf{0.8610} & 0.8578 & 0.8550 & 0.8493 \\
& AVG          & \textbf{0.8391} & 0.8255 & 0.8296 & 0.8210 & \textbf{0.8589} & 0.8519 & 0.8517 & 0.8462 \\
\addlinespace[2pt]

\multirow{3}{*}{Hin$\rightarrow$Tam} 
& Hinglish-Chat& \textbf{0.8786} & 0.8708 & 0.7844 & 0.8576 & 0.8335 & \textbf{0.8400} & 0.7507 & 0.7998 \\
& News-gen     & \textbf{0.8796} & 0.8665 & 0.8175 & 0.8532 & 0.8614 & \textbf{0.8664} & 0.7914 & 0.8288 \\
& AVG          & \textbf{0.8791} & 0.8686 & 0.8010 & 0.8554 & 0.8475 & \textbf{0.8532} & 0.7710 & 0.8143 \\
\addlinespace[2pt]

\multirow{3}{*}{Hin$\rightarrow$Tel} 
& Hinglish-Chat& 0.8375 & \textbf{0.8425} & 0.7921 & 0.8042 & 0.8339 & \textbf{0.8387} & 0.7789 & 0.7945 \\
& News-gen     & \textbf{0.8933} & 0.8893 & 0.8626 & 0.8612 & \textbf{0.8615} & 0.8614 & 0.8294 & 0.8350 \\
& AVG          & 0.8654 & \textbf{0.8659} & 0.8273 & 0.8327 & 0.8477 & \textbf{0.8500} & 0.8041 & 0.8147 \\

\midrule
% =============================================================================
% (b) BaatCheet_Gold_IL2X
% =============================================================================
\multicolumn{10}{c}{\textbf{(b) Human\_Train\_IL2X ($\text{IL}\rightarrow\text{X}$ Direction)}} \\
\midrule

\multirow{4}{*}{Hin$\rightarrow$Eng} 
& DailyTalk    & \textbf{0.8677} & 0.8642 & 0.8415 & 0.8136 & \textbf{0.8537} & 0.8513 & 0.8492 & 0.8337 \\
& MuTual       & 0.8595 & \textbf{0.8735} & 0.8510 & 0.8267 & \textbf{0.8535} & 0.8508 & 0.8501 & 0.8354 \\
& News-gen     & 0.8666 & \textbf{0.8832} & 0.8506 & 0.8268 & \textbf{0.8447} & 0.8425 & 0.8325 & 0.8246 \\
& AVG          & 0.8646 & \textbf{0.8736} & 0.8477 & 0.8223 & \textbf{0.8506} & 0.8482 & 0.8440 & 0.8312 \\
\addlinespace[2pt]

\multirow{4}{*}{Tam$\rightarrow$Eng} 
& DailyTalk    & 0.8331 & \textbf{0.8496} & 0.7968 & 0.7585 & 0.8286 & \textbf{0.8335} & 0.7912 & 0.7656 \\
& MuTual       & 0.8576 & \textbf{0.8751} & 0.8210 & 0.7899 & 0.8357 & \textbf{0.8379} & 0.8029 & 0.7887 \\
& News-gen     & 0.8517 & \textbf{0.8766} & 0.7887 & 0.7923 & 0.8227 & \textbf{0.8258} & 0.7455 & 0.7797 \\
& AVG          & 0.8475 & \textbf{0.8671} & 0.8022 & 0.7802 & 0.8290 & \textbf{0.8324} & 0.7799 & 0.7780 \\
\addlinespace[2pt]

\multirow{4}{*}{Tel$\rightarrow$Eng} 
& DailyTalk    & 0.8571 & \textbf{0.8707} & 0.8341 & 0.7853 & \textbf{0.8268} & 0.8266 & 0.8069 & 0.7492 \\
& MuTual       & 0.8551 & \textbf{0.8652} & 0.8366 & 0.7251 & \textbf{0.8358} & 0.8343 & 0.8154 & 0.7124 \\
& News-gen     & 0.8494 & \textbf{0.8773} & 0.8232 & 0.7872 & 0.8140 & \textbf{0.8143} & 0.7812 & 0.7395 \\
& AVG          & 0.8538 & \textbf{0.8711} & 0.8313 & 0.7659 & \textbf{0.8255} & 0.8251 & 0.8012 & 0.7337 \\
\addlinespace[2pt]

\multirow{3}{*}{Tam$\rightarrow$Hin} 
& Hinglish-Chat& 0.8089 & \textbf{0.8189} & 0.6911 & 0.7553 & 0.7870 & \textbf{0.8026} & 0.7329 & 0.7570 \\
& News-gen     & 0.8170 & \textbf{0.8351} & 0.7367 & 0.7798 & 0.7478 & \textbf{0.7632} & 0.6881 & 0.7047 \\
& AVG          & 0.8130 & \textbf{0.8270} & 0.7139 & 0.7676 & 0.7674 & \textbf{0.7829} & 0.7105 & 0.7308 \\
\addlinespace[2pt]

\multirow{3}{*}{Tel$\rightarrow$Hin} 
& Hinglish-Chat& 0.7837 & \textbf{0.8096} & 0.7091 & 0.6615 & 0.8212 & \textbf{0.8372} & 0.7688 & 0.7044 \\
& News-gen     & 0.8423 & \textbf{0.8692} & 0.8070 & 0.7790 & 0.8363 & \textbf{0.8462} & 0.8153 & 0.7605 \\
& AVG          & 0.8130 & \textbf{0.8394} & 0.7581 & 0.7202 & 0.8288 & \textbf{0.8417} & 0.7921 & 0.7324 \\

\bottomrule
\end{tabular}%
}
\caption{\small Candidate MT evaluation scores on \textit{BaatCheet\_Human\_Train} candidate sets disaggregated by language pair and data source for both (a) $\text{X}\rightarrow\text{IL}$ and (b) $\text{IL}\rightarrow\text{X}$ directions (X are English and Hindi, IL are Hindi, Tamil and Telugu), evaluated using Reference-Based COMET and Reference-Free COMET-QE. Bold values indicate the highest score per row for each evaluation metric block.}
\label{tab:appendix_candidate_mt_combined}
\end{table*}

\subsection{Candidate MT Selection Counts Across Filtering Strategies}
\label{sec:appendix_mt_counts}

Table~\ref{tab:appendix_mt_counts} presents the distribution of candidate MT system selections on \textit{BaatCheet\_Human\_Train} under both filtering strategies: (a)~\textit{Utt-Avg} (dialogue-level engine selection across 510 dialogues) and (b)~\textit{Utt-High} (turn-by-turn optimal candidate selection across 8,427 utterances).

\paragraph{Engine Selection Trends.}
Two primary patterns emerge from the selection distribution:
\begin{itemize}[noitemsep, topsep=2pt, leftmargin=*]
    \item \textbf{Dominance of GMT and GPT-4o-mini:} Across both filtering strategies and directions, \textit{Google Translate} (GMT) and \textit{GPT-4o-mini} collectively account for over 80\% of all selected dialogues and utterances, confirming their status as the highest-quality translation engines in our pool.
    \item \textbf{Directional Asymmetry:} On $\text{X2IL}$ directions ($\text{X}\rightarrow\text{IL}$), GMT is the single most selected engine (accounting for 320/510 dialogues in \textit{Utt-Avg} and 3,506/8,427 turns in \textit{Utt-High} under COMET-QE). Conversely, on $\text{IL2X}$ reverse directions ($\text{IL}\rightarrow\text{X}$), \textit{GPT-4o-mini} becomes the dominant engine (up to 431/510 dialogues in \textit{Utt-Avg} and 4,725/8,427 turns in \textit{Utt-High}), reflecting GPT-4o-mini's strength in generating fluent English and Hindi target outputs.
\end{itemize}

\begin{table*}[th!]
\centering
\small
\setlength{\tabcolsep}{6pt}
\renewcommand{\arraystretch}{1.08}
\resizebox{\textwidth}{!}{%
\begin{tabular}{l cccc cccc}
\toprule
& \multicolumn{4}{c}{\textbf{(a) Utt-Avg Strategy (Dialogue Counts, $N=510$)}} 
& \multicolumn{4}{c}{\textbf{(b) Utt-High Strategy (Utterance Turn Counts, $N=8,427$)}} \\
\cmidrule(lr){2-5} \cmidrule(lr){6-9}
& \multicolumn{2}{c}{\textbf{$\text{X2IL}$ Direction ($\text{X}\rightarrow\text{IL}$)}} 
& \multicolumn{2}{c}{\textbf{$\text{IL2X}$ Direction ($\text{IL}\rightarrow\text{X}$)}} 
& \multicolumn{2}{c}{\textbf{$\text{X2IL}$ Direction ($\text{X}\rightarrow\text{IL}$)}} 
& \multicolumn{2}{c}{\textbf{$\text{IL2X}$ Direction ($\text{IL}\rightarrow\text{X}$)}} \\
\cmidrule(lr){2-3} \cmidrule(lr){4-5} \cmidrule(lr){6-7} \cmidrule(lr){8-9}
\textbf{MT Engine} & \textbf{COMET-QE} & \textbf{COMET} & \textbf{COMET-QE} & \textbf{COMET} & \textbf{COMET-QE} & \textbf{COMET} & \textbf{COMET-QE} & \textbf{COMET} \\
\midrule

GMT          & \textbf{320} & \textbf{320} & 199 & 79 & \textbf{3,506} & \textbf{3,449} & 3,210 & 2,811 \\
GPT-4o-mini  & 154          & 127          & \textbf{294} & \textbf{431} & 2,690 & 2,541 & \textbf{3,720} & \textbf{4,725} \\
IndicTrans2  & 32           & 37           & 16           & 0   & 1,154 & 1,186 & 1,017 & 573   \\
BhashaVerse  & 4            & 26           & 1            & 0   & 1,077 & 1,251 & 480   & 318   \\
\midrule
\textbf{Total} & \textbf{510} & \textbf{510} & \textbf{510} & \textbf{510} & \textbf{8,427} & \textbf{8,427} & \textbf{8,427} & \textbf{8,427} \\

\bottomrule
\end{tabular}%
}
\caption{\small Candidate MT system selection counts on \textit{BaatCheet\_Human\_Train} candidate sets under both (a) \textit{Utt-Avg} (dialogue-level engine selection across 510 dialogues) and (b) \textit{Utt-High} (utterance-level turn selection across 8,427 turns) filtering strategies, evaluated using reference-free COMET-QE and reference-based COMET. Bold text highlights the most selected MT engine per column.}
\label{tab:appendix_mt_counts}
\end{table*}

\subsection{Validation of Filtering Strategies: DOC-COMET and COMTAIL Analysis} \label{sec:appendix_filter_validation}

Table~\ref{tab:appendix_filter_validation} reports the detailed validation of candidate filtering strategies (\textit{Utt-Avg} vs \textit{Utt-High}) evaluated on \textit{BaatCheet\_Human\_Train} candidate outputs using context-aware DOC-COMET and Indic-specialized COMTAIL metrics across all language directions.

% \paragraph{Empirical Superiority of Turn-by-Turn Selection (\textit{Utt-High}).}
When candidate selection is guided by reference-free COMET-QE, \textit{Utt-High} consistently outperforms \textit{Utt-Avg} across both evaluation metrics and all language directions:
\begin{itemize}[noitemsep, topsep=2pt, leftmargin=*]
    \item \textbf{DOC-COMET Alignment:} \textit{Utt-High} achieves higher overall average dialogue scores across both $\text{X2IL}$ ($0.8854$ vs $0.8803$) and $\text{IL2X}$ ($0.8435$ vs $0.8392$), confirming that selecting the best-scoring candidate translation per turn improves multi-turn discourse consistency.
    % \item \textbf{COMTAIL Register Preservation:} On the Indic-trained COMTAIL metric, \textit{Utt-High} achieves substantial gains over \textit{Utt-Avg} for both $\text{X2IL}$ ($0.9078$ vs $0.8967$) and $\text{IL2X}$ ($0.6627$ vs $0.6474$), demonstrating superior preservation of informal Indic register and code-mixed phrasing.
\end{itemize}

\paragraph{Conclusion.}
These results confirm that turn-by-turn COMET-QE selection (\textit{Utt-High}) prevents individual low-quality utterances from degrading dialogue quality, ensuring that no utterance falls below the quality ceiling set by the evaluator. This empirical validation justifies adopting \textbf{Utt-High} as the primary filtering strategy for scaling up the construction of \textit{BaatCheet\_Synthetic}.

\begin{table*}[th!]
\centering
\scriptsize
\setlength{\tabcolsep}{3.5pt}
\renewcommand{\arraystretch}{1.06}
\resizebox{\textwidth}{!}{%
\begin{tabular}{l cccc cccc}
\toprule
& \multicolumn{4}{c}{\textbf{(a) DOC-COMET Metric Validation}} 
& \multicolumn{4}{c}{\textbf{(b) COMTAIL Metric Validation}} \\
\cmidrule(lr){2-5} \cmidrule(lr){6-9}
& \multicolumn{2}{c}{\textbf{Selection via COMET}} & \multicolumn{2}{c}{\textbf{Selection via COMET-QE}} 
& \multicolumn{2}{c}{\textbf{Selection via COMET}} & \multicolumn{2}{c}{\textbf{Selection via COMET-QE}} \\
\cmidrule(lr){2-3} \cmidrule(lr){4-5} \cmidrule(lr){6-7} \cmidrule(lr){8-9}
\textbf{Lang Pair} & \textbf{Utt-Avg} & \textbf{Utt-High} & \textbf{Utt-Avg} & \textbf{Utt-High} & \textbf{Utt-Avg} & \textbf{Utt-High} & \textbf{Utt-Avg} & \textbf{Utt-High} \\
\midrule

\multicolumn{9}{c}{\textbf{(a) 
Human\_Train\_X2IL($\text{X}\rightarrow\text{IL}$)}} \\
\midrule
Eng$\rightarrow$Hin & 0.9049 & 0.9041 & 0.9057 & \textbf{0.9082} & 0.9177 & 0.9189 & 0.9213 & \textbf{0.9263} \\
Eng$\rightarrow$Tam & 0.9012 & 0.9018 & 0.9033 & \textbf{0.9067} & 0.8198 & 0.8209 & 0.8270 & \textbf{0.8466} \\
Eng$\rightarrow$Tel & 0.8932 & 0.8958 & 0.8958 & \textbf{0.9004} & 0.9277 & 0.9305 & 0.9320 & \textbf{0.9371} \\
Hin$\rightarrow$Tam & 0.8272 & 0.8296 & 0.8412 & \textbf{0.8489} & 0.8925 & 0.8946 & 0.8999 & \textbf{0.9151} \\
Hin$\rightarrow$Tel & 0.8538 & 0.8596 & 0.8554 & \textbf{0.8628} & 0.9004 & 0.9011 & 0.9035 & \textbf{0.9140} \\
\midrule
\textbf{Overall AVG} & 0.8761 & 0.8782 & 0.8803 & \textbf{0.8854} & 0.8916 & 0.8932 & 0.8967 & \textbf{0.9078} \\

\midrule
\multicolumn{9}{c}{\textbf{(b) BaatCheet\_Human\_Train\_IL2X ($\text{IL}\rightarrow\text{X}$)}} \\
\midrule
Hin$\rightarrow$Eng & 0.8787 & 0.8789 & 0.8808 & \textbf{0.8840} & 0.5508 & 0.5486 & 0.5532 & \textbf{0.5584} \\
Tam$\rightarrow$Eng & 0.8543 & 0.8538 & 0.8536 & \textbf{0.8594} & 0.5998 & 0.5996 & 0.6000 & \textbf{0.6010} \\
Tel$\rightarrow$Eng & 0.8475 & 0.8469 & 0.8484 & \textbf{0.8522} & 0.5998 & 0.5987 & 0.6015 & \textbf{0.6055} \\
Tam$\rightarrow$Hin & 0.7625 & 0.7626 & 0.7686 & \textbf{0.7733} & 0.6748 & 0.6749 & 0.6844 & \textbf{0.7136} \\
Tel$\rightarrow$Hin & 0.8448 & 0.8433 & 0.8444 & \textbf{0.8487} & 0.7762 & 0.7809 & 0.7976 & \textbf{0.8349} \\
\midrule
\textbf{Overall AVG} & 0.8376 & 0.8371 & 0.8392 & \textbf{0.8435} & 0.6403 & 0.6405 & 0.6474 & \textbf{0.6627} \\

\bottomrule
\end{tabular}%
}
\caption{\small Selection strategy validation on \textit{BaatCheet\_Human\_Train} candidate outputs comparing \textit{Utt-Avg} vs \textit{Utt-High} filtering strategies under both COMET and COMET-QE candidate selection, evaluated using (a) Context-aware DOC-COMET and (b) Indic-specialized COMTAIL metrics across all language directions. Bold text highlights the winning strategy under COMET-QE selection.}
\label{tab:appendix_filter_validation}
\end{table*}

\subsection{Human Binary Preference Evaluation of Curation Strategies}
\label{sec:appendix_binary_preference}
To evaluate human perception of dialogue curation strategies, native expert linguists conducted a binary preference study comparing \textit{Utt-Avg} (dialogue-level engine selection) against \textit{Utt-High} (turn-by-turn optimal selection) on 510 sampled dialogues (255 per translation direction). Annotators evaluated pairwise dialogue outputs on overall naturalness, register consistency, and translation accuracy. Table~\ref{tab:appendix_human_binary_preference} reports the preference counts across both translation axes.
\paragraph{Key Observations.}
\begin{itemize}[noitemsep, topsep=2pt, leftmargin=*]
    \item \textbf{Perceived Naturalness Preference:} Human annotators preferred \textit{Utt-Avg} in 54\% of evaluated dialogues (276/510) compared to 37\% for \textit{Utt-High} (196/510), primarily due to the single-engine stylistic consistency preserved across dialogue turns under \textit{Utt-Avg}.
    \item \textbf{Translation Axis Consistency:} The preference distribution remains highly consistent across both $\text{X2IL}$ (55.9\% \textit{Utt-Avg} vs 40.9\% \textit{Utt-High}) and $\text{IL2X}$ (58.2\% \textit{Utt-Avg} vs 36.8\% \textit{Utt-High}) directions.
    \item \textbf{Strategic Selection:} As analyzed in Section~\ref{Train-Data}, while human evaluators favor \textit{Utt-Avg} for surface naturalness, \textit{Utt-High} guarantees strict turn-level translation accuracy, which is the primary criterion for SFT dataset construction.
\end{itemize}

\begin{figure*}[ht!]
    \centering
    \includegraphics[height=0.4\textheight]{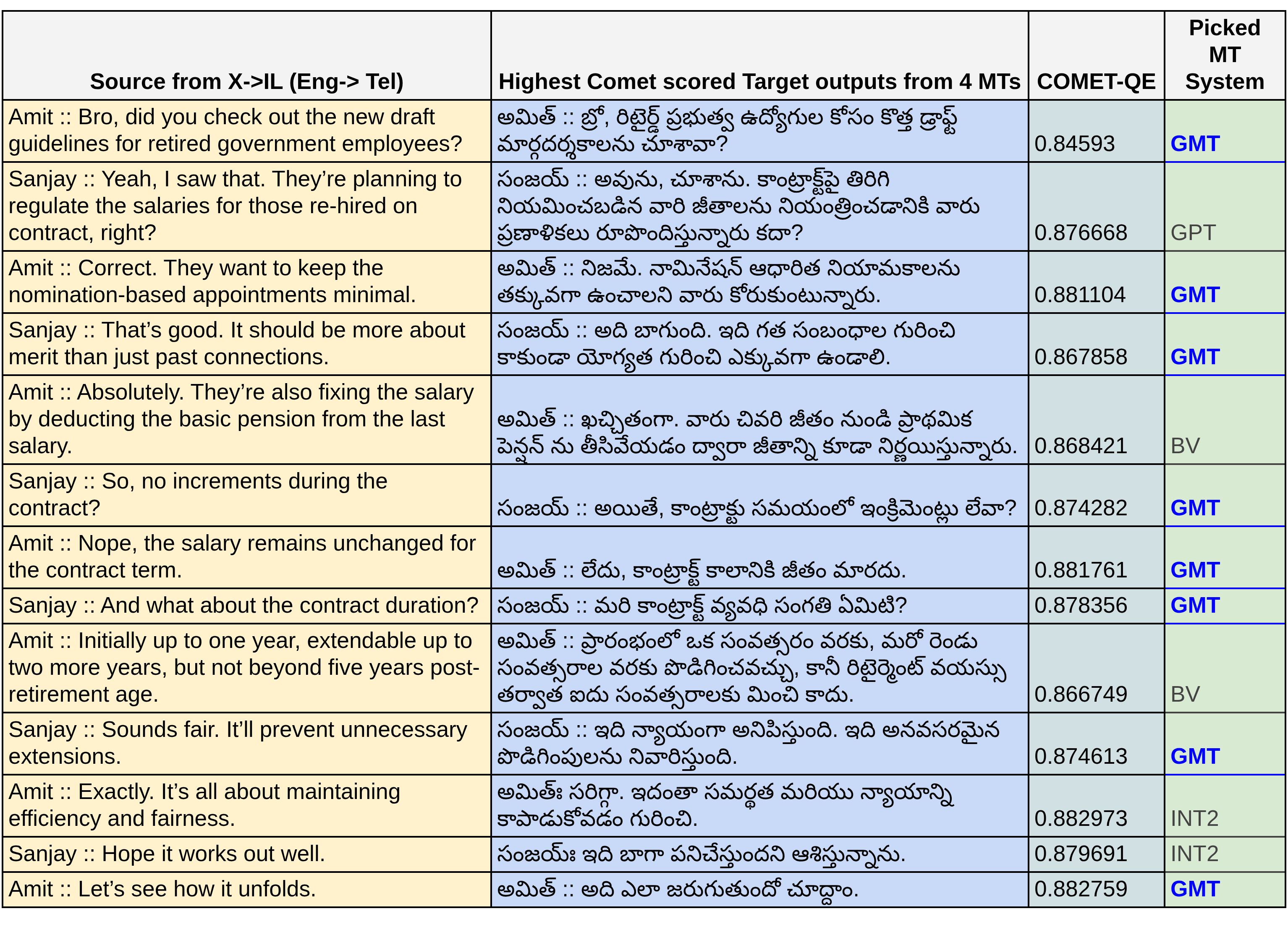}
    \caption{\small An example English$\rightarrow$Telugu dialogue from \textit{BaatCheet\_Human\_Train} 
illustrating the \textit{Utt-High} COMET-QE selection strategy. Each utterance's Telugu translation is independently selected from four candidate MT engines (GMT = Google Translate, GPT = GPT-4o-mini, BV = BhashaVerse, INT2 = IndicTrans2), with the corresponding COMET-QE score shown per turn. This shows that GMT dominance is higher, and the same is reflected in the utt-avg highest score; GMT is 0.869 in the same dialogue.} 
    \label{fig:utt_high_example}
\end{figure*}

Figure~\ref{fig:utt_high_example} provides a concrete illustration of 
\textit{Utt-High} selection on an English$\rightarrow$Telugu dialogue, 
showing how different MT engines are selected at different turns based on 
per-utterance COMET-QE scores. 

\begin{table}[H]
\centering
\small
\renewcommand{\arraystretch}{1.1}
\begin{tabular}{l cc}
\toprule
\textbf{Translation Direction} & \textbf{Utt-Avg} & \textbf{Utt-High}  \\
\midrule
$\text{X2IL}$ Direction ($\text{X}\rightarrow\text{IL}$) & 123 (55.9\%) & 90 (40.9\%) \\
$\text{IL2X}$ Direction ($\text{IL}\rightarrow\text{X}$) & 128 (58.2\%) & 81 (36.8\%)  \\
\midrule
\textbf{Total}                                         & \textbf{251 (57.0\%)} & \textbf{171 (38.9\%)} \\
\bottomrule
\end{tabular}
\caption{\small Human binary preference study results comparing \textit{Utt-Avg} and \textit{Utt-High} filtering strategies across 510 sampled dialogues.}
\label{tab:appendix_human_binary_preference}
\end{table}

% \clearpage
\section{The Fine-Tuning Open-Source LLMs}
\subsection{Finetuning Parameters} \label{parameters}
The table~\ref{tab:hyperparameters} shows the hyperparameters used to fine-tune the selected models. BaatCheet\_Human\_Train configuration we have used 3 epochs to fine-tune the model, and for BaatCheet\_[Syn\_X2IL+HT] configuration we have fine-tuned the model on BaatCheet\_Syn\_X2IL first for 1 epoch and then on BaatCheet\_Human\_Train for 3 epochs.

\begin{table}[H]
\centering
\small
\setlength{\tabcolsep}{3.5pt}
\renewcommand{\arraystretch}{1.1}
\begin{tabular}{lc}
\toprule
\textbf{Hyperparameter} & \textbf{Value} \\
\midrule
Number of epochs & 1 \\
Batch size & 2 \\
Gradient accumulation step & 4 \\
Learning rate & 1e-4 \\
Maximum sequence length & 4096 \\
LoRA rank & 32 \\
LoRA alpha & 64 \\
LoRA dropout & 0.05 \\
init\_lora\_weights & gaussian \\
Target Modules & \makecell{q,k,v,o,gate,\\up,down proj.} \\
Modules to save & lm\_head \\
\bottomrule
\end{tabular}
\caption{\small Training Hyperparameters.}
\label{tab:hyperparameters}
\end{table}

\subsection{Prompt Used for Finetuning the LLMS} \label{train_prompt}

The prompt explicitly instructs the model to produce informal and conversational translations, ensuring that the generated outputs better reflect natural spoken dialogue.

\begin{tcolorbox}[title=Training Prompt Template]  \label{train_prompt}
You are an expert in an informal dialogue machine translation assistant.
Translate the following dialogue from {source\_language} to {target\_language}
into a natural, informal, and colloquial conversational style in
{target\_language}.

Source:
{source\_dialogue}

Target:
{target\_dialogue}
\end{tcolorbox}

\subsection{Detailed Automatic Evaluation Results} \label{Results}

The fine-tuned model's performance results for each dataset configuration and detailed language-pairwise scores are provided in Table~\ref{tab:appendix_comet} for COMET, Table~\ref{tab:doc_comet_results} for DOC-COMET, and Table~\ref{tab:comtail_results} for COMTAIL.

% Detailed auto scores figure

% \begin{figure*}[ht!]
%     \centering
%     \includegraphics[width=\textwidth]{latex/lang_pair_LLMs_scores.png}
%     \caption{COMET, DOC-COMET, and COMTAIL scores on BaatCheet\_Test for all five models across nine training conditions, broken down by language pair. Each row represents one metric; each column represents one language pair. Bars within each condition group represent the five models. Results are utterance-level with reference.} 
%     \label{fig:lang_pair_LLMs_scores}
% \end{figure*}

\begin{table*}[th!]
\centering
\scriptsize
\setlength{\tabcolsep}{2pt}
\begin{tabular}{ll ccccccccc}
\toprule
\textbf{Model} & \textbf{LP} & \textbf{Zero-Shot} & \textbf{Few-Shot} & \textbf{Syn\_X2IL} & \textbf{Syn\_IL2X} & \textbf{HT} & \shortstack{\textbf{Syn\_X2IL+}\\\textbf{HT}} & \shortstack{\textbf{Syn\_IL2X+}\\\textbf{HT}} & \shortstack{\textbf{Syn\_Both}\\\textbf{+HT}} & \shortstack{\textbf{Syn\_Both}\\\textbf{+3*HT}} \\
\midrule
\multirow{6}{*}{Llama-3B} 
& Eng-Hin & 0.496 & 0.558 & \best{0.751} & 0.748 & 0.748 & 0.747 & 0.736 & 0.712 & 0.604 \\
& Eng-Tam & 0.539 & 0.572 & \best{0.775} & 0.699 & 0.672 & 0.733 & 0.670 & 0.769 & 0.727 \\
& Eng-Tel & 0.523 & 0.556 & 0.784 & 0.694 & 0.699 & 0.711 & 0.694 & \best{0.786} & 0.759 \\
& Hin-Tam & 0.567 & 0.642 & \best{0.790} & 0.753 & 0.723 & 0.764 & 0.730 & 0.778 & 0.786 \\
& Hin-Tel & 0.533 & 0.605 & \best{0.801} & 0.746 & 0.728 & 0.751 & 0.708 & 0.779 & 0.760 \\
\cmidrule{2-11}
& \textbf{AVGs} & 0.532 & 0.586 & \best{0.780} & 0.728 & 0.714 & 0.741 & 0.708 & 0.765 & 0.727 \\
\midrule
\multirow{6}{*}{Llama-8B} 
& Eng-Hin & 0.633 & 0.637 & 0.792 & 0.770 & 0.754 & 0.745 & 0.752 & 0.794 & \best{0.802} \\
& Eng-Tam & 0.701 & 0.680 & 0.800 & 0.717 & 0.790 & 0.702 & 0.697 & 0.792 & \best{0.841} \\
& Eng-Tel & 0.723 & 0.728 & 0.812 & 0.706 & 0.778 & 0.700 & 0.696 & 0.796 & \best{0.857} \\
& Hin-Tam & 0.741 & 0.698 & 0.804 & 0.757 & 0.799 & 0.733 & 0.742 & 0.788 & \best{0.857} \\
& Hin-Tel & 0.759 & 0.698 & 0.804 & 0.760 & 0.815 & 0.719 & 0.718 & 0.795 & \best{0.874} \\
\cmidrule{2-11}
& \textbf{AVGs} & 0.711 & 0.688 & 0.802 & 0.742 & 0.787 & 0.720 & 0.721 & 0.793 & \best{0.846} \\
\midrule
\multirow{6}{*}{Gemma} 
& Eng-Hin & 0.430 & 0.742 & 0.765 & 0.727 & \best{0.800} & 0.782 & 0.792 & 0.683 & 0.606 \\
& Eng-Tam & 0.480 & 0.602 & 0.795 & 0.630 & \best{0.843} & 0.827 & 0.833 & 0.694 & 0.555 \\
& Eng-Tel & 0.438 & 0.557 & 0.821 & 0.774 & \best{0.839} & 0.819 & 0.825 & 0.777 & 0.547 \\
& Hin-Tam & 0.587 & 0.688 & 0.808 & 0.645 & \best{0.834} & 0.829 & 0.833 & 0.736 & 0.698 \\
& Hin-Tel & 0.578 & 0.616 & \best{0.850} & 0.772 & 0.845 & 0.838 & 0.829 & 0.808 & 0.639 \\
\cmidrule{2-11}
& \textbf{AVGs} & 0.502 & 0.641 & 0.808 & 0.710 & \best{0.832} & 0.819 & 0.822 & 0.740 & 0.609 \\
\midrule
\multirow{6}{*}{Sarvam} 
& Eng-Hin & 0.590 & 0.464 & 0.811 & 0.779 & \best{0.825} & 0.822 & 0.821 & 0.805 & 0.788 \\
& Eng-Tam & 0.679 & 0.527 & 0.862 & 0.819 & 0.866 & \best{0.868} & 0.851 & 0.830 & 0.854 \\
& Eng-Tel & 0.691 & 0.463 & \best{0.867} & 0.771 & 0.862 & 0.862 & 0.848 & 0.862 & 0.832 \\
& Hin-Tam & 0.537 & 0.517 & \best{0.871} & 0.818 & 0.747 & 0.865 & 0.865 & 0.863 & 0.857 \\
& Hin-Tel & 0.587 & 0.457 & 0.665 & 0.842 & 0.860 & 0.870 & 0.864 & \best{0.876} & 0.875 \\
\cmidrule{2-11}
& \textbf{AVGs} & 0.617 & 0.486 & 0.815 & 0.806 & 0.832 & \best{0.857} & 0.850 & 0.847 & 0.841 \\
\midrule
\multirow{6}{*}{Qwen} 
& Eng-Hin & 0.717 & 0.638 & \best{0.792} & 0.750 & 0.765 & 0.770 & 0.727 & 0.713 & 0.790 \\
& Eng-Tam & 0.630 & 0.634 & 0.841 & 0.718 & 0.779 & 0.815 & 0.719 & 0.749 & \best{0.849} \\
& Eng-Tel & 0.653 & 0.638 & \best{0.846} & 0.716 & 0.784 & 0.794 & 0.722 & 0.826 & 0.838 \\
& Hin-Tam & 0.694 & 0.690 & \best{0.852} & 0.746 & 0.827 & 0.829 & 0.766 & 0.759 & 0.843 \\
& Hin-Tel & 0.705 & 0.702 & \best{0.855} & 0.757 & 0.824 & 0.823 & 0.751 & 0.801 & 0.824 \\
\cmidrule{2-11}
& \textbf{AVGs} & 0.680 & 0.660 & \best{0.837} & 0.737 & 0.796 & 0.806 & 0.737 & 0.770 & 0.829 \\
\bottomrule
\end{tabular}
\caption{Utterance-level COMET scores on BaatCheet\_Test. Shaded cells represent the best performance per language pair. \texttt{Syn\_Both} denotes $\text{Syn\_[X2IL+IL2X]}$ and \texttt{HT} denotes $\text{Human\_Train}$}
\label{tab:appendix_comet}
\end{table*}

\begin{table*}[th!]
\centering
\scriptsize
\setlength{\tabcolsep}{2.6pt}
\begin{tabular}{ll ccccccccc}
\toprule
\textbf{Model} & \textbf{LP} & \textbf{Zero-Shot} & \textbf{Few-Shot} & \textbf{Syn\_X2IL} & \textbf{Syn\_IL2X} & \textbf{HT} & \shortstack{\textbf{Syn\_X2IL+}\\\textbf{HT}} & \shortstack{\textbf{Syn\_IL2X+}\\\textbf{HT}} & \shortstack{\textbf{Syn\_Both}\\\textbf{+HT}} & \shortstack{\textbf{Synt
\_Both}\\\textbf{+3*HT}} \\
\midrule
\multirow{6}{*}{Llama-3B} 
& Eng-Hin & \best{0.675} & 0.495 & 0.710 & 0.700 & 0.705 & 0.705 & 0.688 & 0.661 & 0.507 \\
& Eng-Tam & 0.556 & 0.552 & 0.682 & 0.577 & 0.587 & \best{0.675} & 0.584 & 0.688 & 0.659 \\
& Eng-Tel & 0.590 & 0.525 & \best{0.713} & 0.581 & 0.601 & 0.650 & 0.608 & 0.710 & 0.699 \\
& Hin-Tam & 0.499 & 0.531 & 0.555 & 0.602 & 0.588 & 0.590 & 0.593 & 0.617 & \best{0.729} \\
& Hin-Tel & 0.523 & 0.474 & 0.572 & 0.597 & 0.563 & 0.589 & 0.578 & 0.635 & \best{0.700} \\
\cmidrule{2-11}
& \textbf{AVGs} & 0.568 & 0.515 & 0.647 & 0.611 & 0.609 & 0.642 & 0.610 & \best{0.662} & 0.658 \\
\midrule
\multirow{6}{*}{Llama-8B} 
& Eng-Hin & 0.520 & 0.567 & 0.747 & 0.734 & 0.714 & 0.702 & 0.709 & 0.753 & \best{0.764} \\
& Eng-Tam & 0.640 & 0.601 & 0.740 & 0.611 & 0.753 & 0.644 & 0.639 & 0.727 & \best{0.785} \\
& Eng-Tel & 0.679 & 0.654 & 0.752 & 0.623 & 0.739 & 0.644 & 0.634 & 0.734 & \best{0.803} \\
& Hin-Tam & 0.688 & 0.620 & 0.597 & 0.620 & 0.771 & 0.558 & 0.590 & 0.588 & \best{0.802} \\
& Hin-Tel & 0.715 & 0.593 & 0.614 & 0.621 & 0.753 & 0.586 & 0.591 & 0.597 & \best{0.810} \\
\cmidrule{2-11}
& \textbf{AVGs} & 0.648 & 0.607 & 0.690 & 0.642 & 0.746 & 0.626 & 0.632 & 0.680 & \best{0.793} \\
\midrule
\multirow{6}{*}{Gemma} 
& Eng-Hin & 0.352 & 0.259 & 0.717 & 0.681 & \best{0.756} & 0.737 & 0.749 & 0.629 & 0.620 \\
& Eng-Tam & 0.376 & 0.315 & 0.737 & 0.477 & \best{0.793} & 0.776 & 0.781 & 0.595 & 0.566 \\
& Eng-Tel & 0.331 & 0.280 & 0.781 & 0.704 & \best{0.792} & 0.771 & 0.779 & 0.710 & 0.488 \\
& Hin-Tam & 0.520 & 0.524 & 0.741 & 0.558 & 0.775 & 0.778 & \best{0.783} & 0.663 & 0.626 \\
& Hin-Tel & 0.494 & 0.500 & \best{0.780} & 0.751 & 0.769 & 0.775 & 0.761 & 0.736 & 0.586 \\
\cmidrule{2-11}
& \textbf{AVGs} & 0.415 & 0.376 & 0.751 & 0.634 & \best{0.777} & 0.767 & 0.770 & 0.667 & 0.577 \\
\midrule
\multirow{6}{*}{Sarvam} 
& Eng-Hin & 0.604 & 0.333 & 0.766 & 0.747 & 0.782 & 0.776 & \best{0.782} & 0.759 & 0.759 \\
& Eng-Tam & 0.475 & 0.440 & 0.809 & 0.746 & \best{0.822} & 0.819 & 0.816 & 0.772 & 0.796 \\
& Eng-Tel & 0.644 & 0.384 & 0.815 & 0.791 & \best{0.817} & 0.815 & 0.812 & 0.809 & 0.807 \\
& Hin-Tam & 0.356 & 0.305 & \best{0.818} & 0.743 & 0.802 & 0.812 & 0.817 & 0.808 & 0.808 \\
& Hin-Tel & 0.398 & 0.283 & 0.596 & 0.776 & 0.789 & 0.800 & 0.794 & \best{0.810} & 0.808 \\
\cmidrule{2-11}
& \textbf{AVGs} & 0.495 & 0.349 & 0.760 & 0.761 & 0.802 & \best{0.804} & \best{0.804} & 0.796 & 0.792 \\
\midrule
\multirow{6}{*}{Qwen} 
& Eng-Hin & 0.683 & 0.617 & \best{0.750} & 0.710 & 0.721 & 0.729 & 0.698 & 0.732 & 0.745 \\
& Eng-Tam & 0.522 & 0.552 & 0.783 & 0.600 & 0.722 & 0.760 & 0.661 & 0.797 & \best{0.797} \\
& Eng-Tel & 0.559 & 0.552 & \best{0.798} & 0.617 & 0.718 & 0.734 & 0.677 & 0.789 & 0.789 \\
& Hin-Tam & 0.600 & 0.591 & 0.798 & 0.622 & 0.785 & 0.778 & 0.677 & \best{0.799} & 0.799 \\
& Hin-Tel & 0.640 & 0.598 & \best{0.795} & 0.620 & 0.759 & 0.744 & 0.645 & 0.793 & 0.793 \\
\cmidrule{2-11}
& \textbf{AVGs} & 0.601 & 0.582 & 0.785 & 0.634 & 0.741 & 0.749 & 0.672 & 0.784 & \best{0.784} \\
\bottomrule
\end{tabular}
\caption{DOC-COMET on BaatCheet\_Test. Shaded cells represent the best performance per language pair. \texttt{Syn\_Both} denotes $\text{Syn\_[X2IL+IL2X]}$ and \texttt{HT} denotes $\text{Human\_Train}$}
\label{tab:doc_comet_results}
\end{table*}

\begin{table*}[th!]
\centering
\scriptsize
\begin{tabular}{ll ccccccccc}
\toprule
\textbf{Model} & \textbf{LP} & \textbf{Zero-Shot} & \textbf{Few-Shot} & \textbf{Syn\_X2IL} & \textbf{Syn\_IL2X} & \textbf{HT} & \shortstack{\textbf{Syn\_X2IL+}\\\textbf{HT}} & \shortstack{\textbf{Syn\_IL2X+}\\\textbf{HT}} & \shortstack{\textbf{Synt
\_Both}\\\textbf{+HT}} & \shortstack{\textbf{Syn\_Both}\\\textbf{+3*HT}} \\
\midrule
\multirow{6}{*}{Llama-3B} 
& Eng-Hin & 0.394 & 0.498 & \best{0.755} & 0.715 & 0.723 & 0.717 & 0.713 & 0.698 & 0.555 \\
& Eng-Tam & 0.417 & 0.434 & \best{0.615} & 0.524 & 0.513 & 0.558 & 0.506 & 0.610 & 0.553 \\
& Eng-Tel & 0.301 & 0.315 & 0.700 & 0.524 & 0.515 & 0.536 & 0.497 & \best{0.701} & 0.658 \\
& Hin-Tam & 0.390 & 0.430 & \best{0.621} & 0.549 & 0.529 & 0.567 & 0.535 & 0.609 & 0.604 \\
& Hin-Tel & 0.306 & 0.408 & \best{0.716} & 0.606 & 0.581 & 0.621 & 0.553 & 0.681 & 0.648 \\
\cmidrule{2-11}
& \textbf{AVGs} & 0.362 & 0.417 & \best{0.681} & 0.584 & 0.572 & 0.600 & 0.561 & 0.660 & 0.604 \\
\midrule
\multirow{6}{*}{Llama-8B} 
& Eng-Hin & 0.480 & 0.601 & \best{0.806} & 0.736 & 0.735 & 0.724 & 0.728 & 0.804 & 0.805 \\
& Eng-Tam & 0.524 & 0.534 & 0.635 & 0.528 & 0.586 & 0.534 & 0.534 & 0.624 & \best{0.651} \\
& Eng-Tel & 0.535 & 0.588 & 0.745 & 0.534 & 0.617 & 0.507 & 0.494 & 0.718 & \best{0.814} \\
& Hin-Tam & 0.512 & 0.484 & 0.637 & 0.548 & 0.588 & 0.518 & 0.531 & 0.622 & \best{0.694} \\
& Hin-Tel & 0.564 & 0.539 & 0.713 & 0.614 & 0.715 & 0.573 & 0.564 & 0.705 & \best{0.818} \\
\cmidrule{2-11}
& \textbf{AVGs} & 0.523 & 0.549 & 0.707 & 0.592 & 0.648 & 0.571 & 0.570 & 0.694 & \best{0.756} \\
\midrule
\multirow{6}{*}{Gemma} 
& Eng-Hin & 0.261 & 0.681 & 0.755 & 0.634 & \best{0.789} & 0.764 & 0.779 & 0.608 & 0.539 \\
& Eng-Tam & 0.217 & 0.423 & 0.608 & 0.406 & \best{0.653} & 0.634 & 0.635 & 0.513 & 0.411 \\
& Eng-Tel & 0.138 & 0.397 & 0.763 & 0.653 & \best{0.764} & 0.718 & 0.719 & 0.665 & 0.341 \\
& Hin-Tam & 0.307 & 0.465 & 0.603 & 0.463 & \best{0.649} & 0.637 & 0.635 & 0.509 & 0.492 \\
& Hin-Tel & 0.355 & 0.429 & \best{0.780} & 0.671 & 0.768 & 0.751 & 0.725 & 0.711 & 0.492 \\
\cmidrule{2-11}
& \textbf{AVGs} & 0.256 & 0.479 & 0.702 & 0.565 & \best{0.724} & 0.701 & 0.699 & 0.601 & 0.455 \\
\midrule
\multirow{6}{*}{Sarvam} 
& Eng-Hin & 0.532 & 0.315 & \best{0.826} & 0.770 & 0.825 & 0.821 & 0.817 & 0.820 & 0.800 \\
& Eng-Tam & 0.514 & 0.389 & \best{0.688} & 0.607 & 0.677 & 0.671 & 0.654 & 0.665 & 0.676 \\
& Eng-Tel & 0.579 & 0.290 & \best{0.839} & 0.707 & 0.800 & 0.803 & 0.781 & 0.829 & 0.801 \\
& Hin-Tam & 0.364 & 0.360 & \best{0.707} & 0.615 & 0.581 & 0.688 & 0.685 & 0.696 & 0.694 \\
& Hin-Tel & 0.373 & 0.209 & 0.541 & 0.749 & 0.780 & 0.794 & 0.784 & \best{0.821} & \best{0.821} \\
\cmidrule{2-11}
& \textbf{AVGs} & 0.472 & 0.313 & 0.720 & 0.690 & 0.733 & 0.755 & 0.744 & \best{0.766} & 0.758 \\
\midrule
\multirow{6}{*}{Qwen} 
& Eng-Hin & 0.698 & 0.615 & \best{0.807} & 0.731 & 0.752 & 0.759 & 0.708 & 0.720 & 0.795 \\
& Eng-Tam & 0.496 & 0.494 & \best{0.666} & 0.527 & 0.591 & 0.615 & 0.541 & 0.584 & 0.663 \\
& Eng-Tel & 0.510 & 0.467 & \best{0.802} & 0.542 & 0.646 & 0.667 & 0.550 & 0.766 & 0.787 \\
& Hin-Tam & 0.534 & 0.512 & \best{0.687} & 0.533 & 0.641 & 0.638 & 0.565 & 0.599 & 0.679 \\
& Hin-Tel & 0.586 & 0.566 & \best{0.791} & 0.619 & 0.730 & 0.727 & 0.614 & 0.714 & 0.762 \\
\cmidrule{2-11}
& \textbf{AVGs} & 0.565 & 0.531 & \best{0.751} & 0.590 & 0.672 & 0.681 & 0.596 & 0.676 & 0.737 \\
\bottomrule
\end{tabular}
\caption{COMTAIL Scores on BaatCheet\_Test. Shaded cells represent the best performance per language pair. \texttt{Syn\_Both} denotes $\text{Syn\_[X2IL+IL2X]}$ and \texttt{HT} denotes $\text{Human\_Train}$}
\label{tab:comtail_results}
\end{table*}

\section{SQM Guided DA Evaluation}\label{appendix_normalization}

We conduct a fine-grained evaluation on these top two configurations of Sarvam-Translate using a Scalar Quality Metric (SQM)-guided Direct Assessment (DA) protocol on a continuous 0 to 100 scale. From each setup, we sample test dialogues across all language directions: 30 dialogues for each English-source pair ($\text{Eng}\rightarrow\text{Hin/Tam/Tel}$) and 20 dialogues for the Hindi-source pair ($\text{Hin}\rightarrow\text{Tel}$). In total, 1,100 dialogues (900 English-source and 200 Hindi-source) are independently evaluated by expert native-speaking linguists and two SOTA LLM judges:

\subsection{LLM Prompting and Rubrics} \label{LLM_judges}
The LLM judges were provided with a 7-point rating rubric injected into the system prompt to calibrate the 0--100 scale. We utilized few-shot examples to align the scoring distribution. The full prompt used for both dialogue level and utterance-level evaluation is detailed in Figure \ref{fig:dialogue_prompt_template} and Figure~\ref{fig:utterance_prompt}.

\begin{figure*}[ht]
\centering
\begin{tcolorbox}[
    colback=gray!5, 
    colframe=black, 
    title= Dialogue Level DA+SQM-based Scoring Prompt Template for LLM-as-Judge
    fonttitle=\bfseries,
    arc=0mm,
    left=5pt,
    right=5pt
]
\small
\textbf{System Prompt} \\
\textit{You are an expert evaluator for dialogue translation and have the ability to evaluate the translated dialogues on the basis of naturalness, informality and cohesion and coherence. You will evaluate the provided Target Dialogue (translated from the Source Dialogue). You need to evaluate the entire Target Dialogue on three factors: Naturalness, Informality, and Cohesion \& Coherence by understanding the context of the dialogue.}

\medskip
\textbf{Few-shot Injection} \\
\fbox{\parbox{0.95\textwidth}{\texttt{You have been provided with some examples of human evaluations below... Do not copy the scores and patterns from given example.}}}

\medskip
\textbf{Evaluation Rubrics} \\
\begin{tabularx}{\textwidth}{X X X}
\toprule
\textbf{Naturalness} & \textbf{Informality} & \textbf{Cohesion \& Coherence} \\
\midrule
\textbf{00-15:} Ungrammatical/Unnatural & \textbf{00-15:} Highly Formal & \textbf{00-15:} Incoherent/Incohesive \\
\textbf{16-30:} Ungram. but natural-ish & \textbf{16-30:} Very Formal & \textbf{16-30:} Coherent but Incohesive \\
\textbf{31-45:} Minor errors, natural-ish & \textbf{31-45:} Moderately Formal & \textbf{31-45:} Somew. coherent/cohesive \\
\textbf{46-60:} Grammatical but unnatural & \textbf{46-60:} Neutral & \textbf{46-60:} Coherent/partially cohesive \\
\textbf{61-75:} Grammatical/somew. natural & \textbf{61-75:} Slightly Informal & \textbf{61-75:} Somew. coherent/cohesive \\
\textbf{76-90:} Grammatical/natural (minor flaws) & \textbf{76-90:} Very Informal & \textbf{76-90:} Strong Coherence/Cohesion \\
\textbf{91-100:} Grammatical and Natural & \textbf{91-100:} Extremely Casual & \textbf{91-100:} Fully Coherent/Cohesive \\
\bottomrule
\end{tabularx}

\medskip
\textbf{Constraint:} \textit{DA Score should be on a continuous scale of 0 to 100 representing the overall quality for that factor. Provide precise, granular scores. Do not round to the nearest 5 or 10.}

\medskip
\textbf{User Input} \\
\begin{tabularx}{\textwidth}{|X|X|}
\hline
\textbf{Source Dialogue} & \textbf{Target Dialogue (MT)} \\
\hline
\texttt{\{source\_text\}} & \texttt{\{mt\_text\}} \\
\hline
\end{tabularx}

\medskip
\textbf{Response Instruction:} \texttt{Please evaluate the Target Dialogue. Don't provide any other explanations. Give output in the required format.}
\end{tcolorbox}
\caption{The prompt interface used for LLM-as-a-judge evaluation for Dialogue Level Evaluation. Text in brackets denotes dynamic variables.}
\label{fig:dialogue_prompt_template}
\end{figure*}

\begin{figure*}[ht!]
\centering
\begin{tcolorbox}[
    colback=gray!2, 
    colframe=black, 
    title= Utterance-Level DA+SQM-based Scoring Prompt Template for LLM-as-Judge 
    fonttitle=\bfseries,
    arc=0mm,
    left=8pt,
    right=8pt
]
\small
\textbf{System Prompt} \\
\textit{You are an expert evaluator for dialogue translation. You have the ability to evaluate the translated utterances. You will evaluate the provided Target utterances corresponding to Source utterances. You need to evaluate MULTIPLE utterances simultaneously. For each utterance, provide a single DA score (0-100) based on the rubric.}

\medskip
\textbf{Few-shot Injection (Optional)} \\
\fbox{\parbox{0.96\textwidth}{\texttt{You have been provided with some examples of human evaluations below... Use this understanding to calibrate your own scoring for the new utterances.}}}

\medskip
\textbf{Utterance-Level Scoring Rubric (0-100)} \\
\begin{tabularx}{\textwidth}{l|X}
\toprule
\textbf{Score Range} & \textbf{Evaluation Criteria} \\
\midrule
\textbf{00--15} & \textbf{Completely unusable:} Nonsensical/total mistranslation; broken grammar; literal word-for-word translation; original meaning cannot be recovered. \\
\midrule
\textbf{16--30} & \textbf{Very poor translation:} Major errors in meaning/structure; important info added/omitted; extremely awkward word order; general topic only guessable. \\
\midrule
\textbf{31--45} & \textbf{Meaning partially preserved:} Recognizable source meaning but several errors remain; grammatical/lexical issues affect clarity; intent may be distorted. \\
\midrule
\textbf{46--60} & \textbf{Meaning correct but unnatural:} Core meaning preserved; grammatically acceptable but sounds literal/stiff; uses formal words in conversational context. \\
\midrule
\textbf{61--75} & \textbf{Mostly accurate/somewhat natural:} Message correct; minor awkward phrasing or slightly stiff word order; tone and intent generally preserved. \\
\midrule
\textbf{76--90} & \textbf{Accurate and natural:} Core message/tone clearly preserved; sounds natural and conversational; minor stylistic imperfections that do not affect fluency. \\
\midrule
\textbf{91--100} & \textbf{Accurate and fully Natural:} Meaning, tone, and intent perfectly preserved; native-like phrasing and word order; correct honorifics and conversational markers. \\
\bottomrule
\end{tabularx}

\medskip
\textbf{Constraints:} \textit{DA Score should be on a continuous scale of 0 to 100. Provide precise, granular scores. Do not round to the nearest 5 or 10. Provide exactly the same number of scores as the number of utterances provided.}

\medskip
\textbf{User Input Prompt} \\
\begin{tcolorbox}[colback=white, colframe=gray!30, arc=0mm]
\texttt{Here are \{num\_utterances\} utterances to evaluate:} \\
\texttt{\{paired\_utterances\_text\}} \\
\texttt{Please evaluate the Target utterances and return a list of scores.}
\end{tcolorbox}

\end{tcolorbox}
\caption{The prompt used for utterance-level evaluation. The dynamic variable \texttt{\{paired\_utterances\_text\}} contains the list of source and target sentence pairs.}
\label{fig:utterance_prompt}
\end{figure*}

% \paragraph{LLM-as-Judge Evaluation Results} \label{LLM_judges}

% \paragraph{Dialogue-level LLM-as-Judge Performance}
% Detailed dialogue-level evaluation scores across four language pairs, as assessed by GPT and Gemini judges. Metrics include Naturalness (\textbf{Nat.}), Informality (\textbf{Inf.}), and Cohesion \& Coherence (\textbf{Coh.}). Each section compares two distinct fine-tuning experiments per model. \textbf{AVG} denotes the mean score across all language pairs for a specific configuration. $H$ refers to Gold and $Syn\_X2IL$ refers to Synthetic\_I data.
% \label{tab:dialogue_full_appendix}

\subsection{Human Direct Assessment} \label{Human_DA}

\begin{figure*}[ht!]
\centering
\includegraphics[height=0.25\textheight]{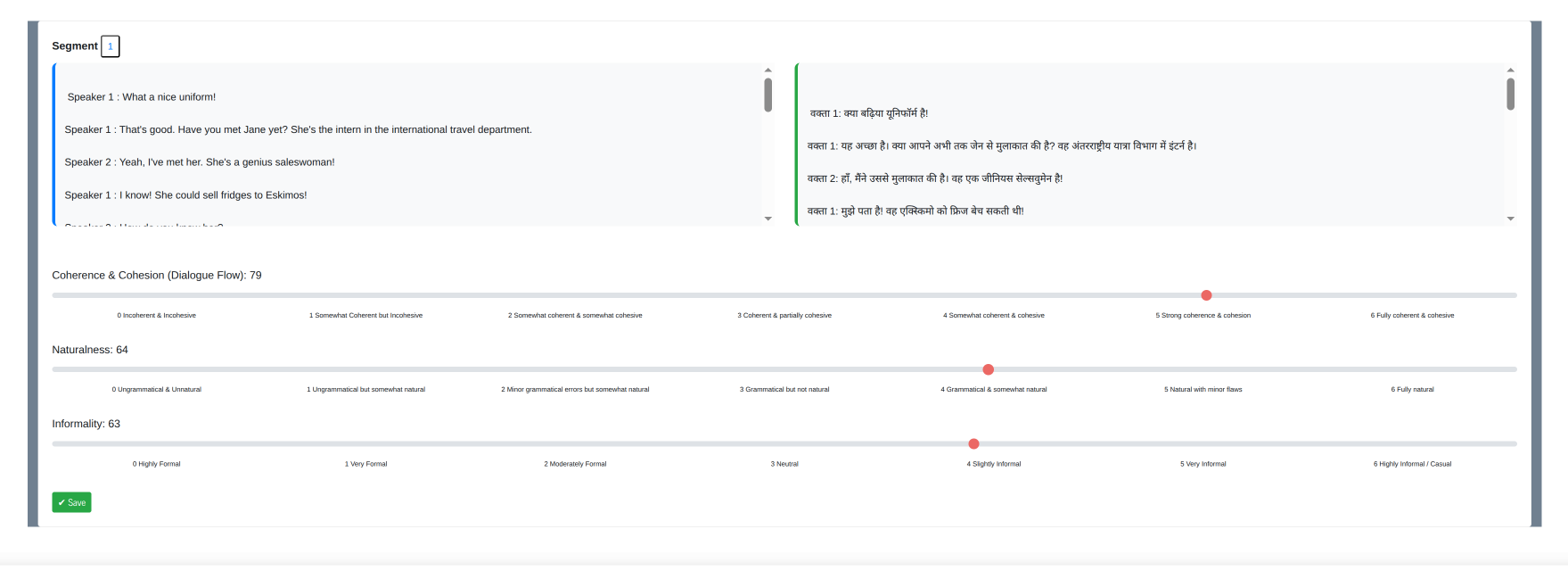}
\caption{\small Dialogue-Level Annotation Platform}
\label{fig:dialogue_tool}
\end{figure*}

\begin{figure*}[ht!]
\centering
\includegraphics[height=0.25\textheight]{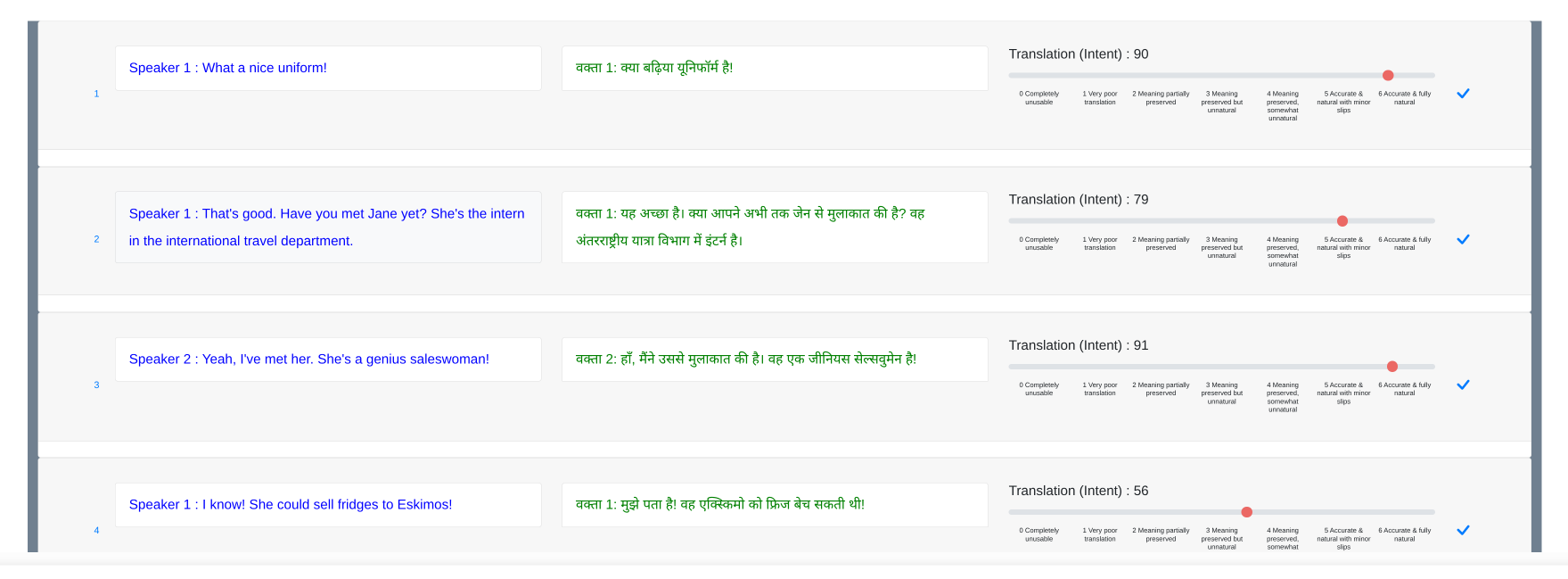}
\caption{\small Utterence-Level Annotation Platform}
\label{fig:utterance_tool}
\end{figure*}

\paragraph{Annotator Selection and Screening}
To ensure high-fidelity annotations, we implemented a strict crowd-annotator screening protocol. Prospective bilingual candidates completed a 12-item source comprehension test with multiple-choice questions evaluating grammar, colloquial register, and translation accuracy. Only candidates scoring greater than or equal to 9/12 were selected for the evaluation task.
Human evaluation was conducted on the \textit{posteditme} platform\footnote{https://posteditme.revanai.in/}. Annotations were performed over three iterations by professional linguists. Annotation tool samples are shown in Figures~\ref{fig:dialogue_tool} and ~\ref{fig:utterance_tool}
\paragraph{Score Aggregation}
To eliminate evaluator-specific rating leniency and scale biases, raw scores are standard-normalized ($Z$-scored) per specific evaluator-language pair group, clipped to $[-3.0, 3.0]$ to suppress extreme outlier variance, and linearly rescaled to an intuitive positive $0\text{--}100$ range.

\begin{figure*}[ht!]
\centering
\includegraphics[height=0.38\textheight]{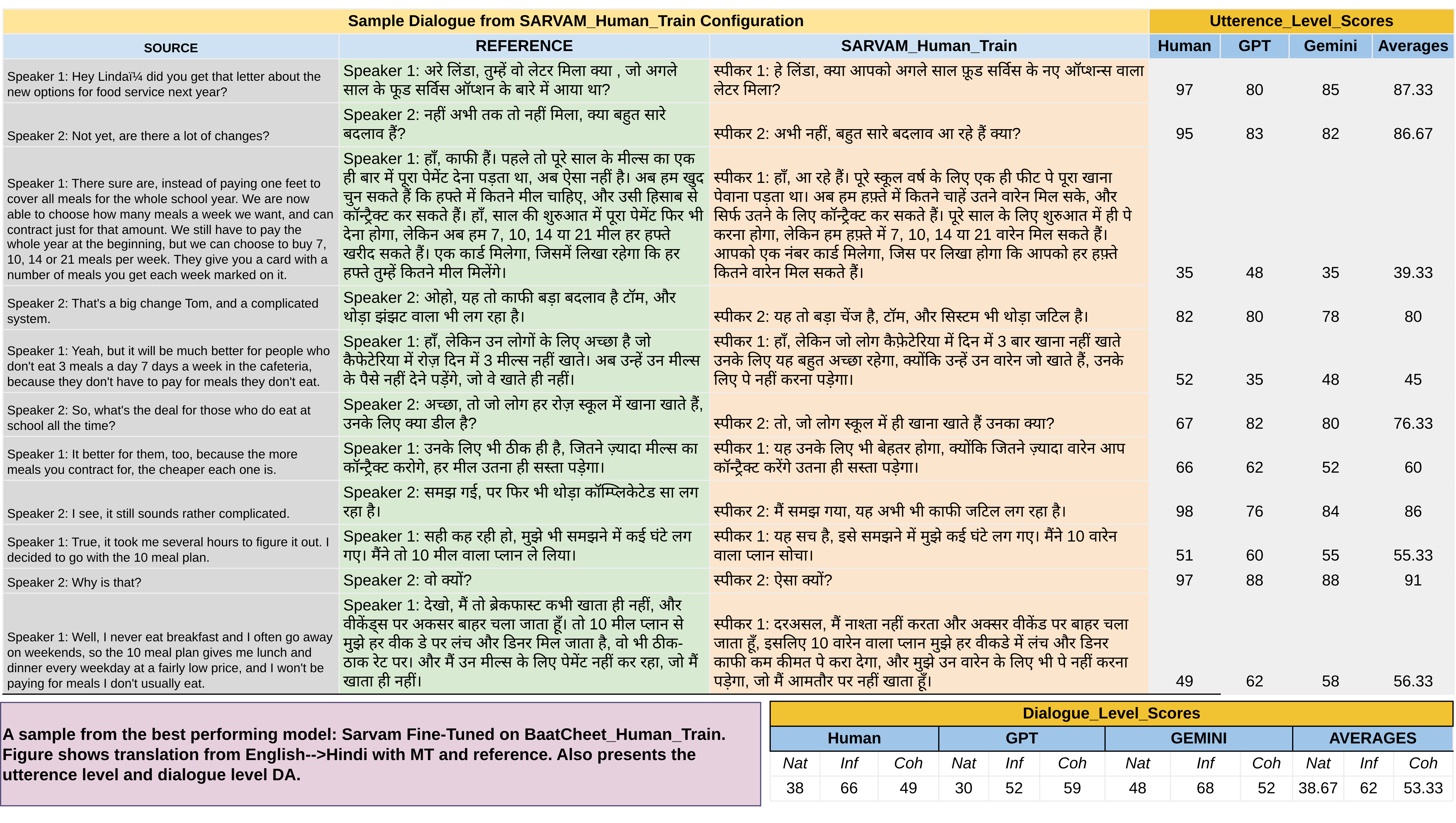}
\caption{\small A sample dialogue in our error analysis suggests that there are still formal-style dialogues. While the baseline semantics are preserved, the translations are only somewhat natural. This finding supports our conclusion that synthetic data alone is insufficient for capturing deep conversational nuances, reinforcing the necessity for larger-scale, human-curated informal dialogue datasets.}
\label{fig:sarvam_dialogue_sample}
\end{figure*}

\begin{table*}[th!]
\centering
\small
\setlength{\tabcolsep}{6pt}
\renewcommand{\arraystretch}{1.08}
\begin{tabular}{l ccccc}
\toprule
\textbf{7-Scale SQM Rubric} & \textbf{Eng--Hin} & \textbf{Eng--Tam} & \textbf{Eng--Tel} & \textbf{Hin--Tam} & \textbf{Hin--Tel} \\
\midrule

% --- (a) UTTERANCE LEVEL ---
\multicolumn{6}{l}{\textbf{(a) Utterance-Level Accuracy}} \\
\midrule
00--15: Completely unusable                  & 0  & 0  & 0  & 0  & 0  \\
16--30: Very poor translation                & 0  & 0  & 0  & 0  & 0  \\
31--45: Meaning partially preserved           & 0  & 0  & 0  & 0  & 0  \\
46--60: Meaning correct but unnatural        & 0  & 1  & 0  & 1  & 1  \\
61--75: Mostly accurate and somewhat natural & 6  & \textbf{25} & \textbf{23} & \textbf{13} & \textbf{11} \\
76--90: Accurate and natural (minor slips)   & \textbf{24} & 4  & 7  & 6  & 8  \\
91--100: Accurate and fully Natural          & 0  & 0  & 0  & 0  & 0  \\
\midrule

% --- (b) DIALOGUE LEVEL: NATURALNESS ---
\multicolumn{6}{l}{\textbf{(b) Dialogue-Level: Naturalness}} \\
\midrule
00--15: Ungrammatical and Unnatural          & 0  & 0  & 0  & 0  & 0  \\
16--30: Ungrammatical but somewhat natural   & 0  & 0  & 0  & 0  & 0  \\
31--45: Minor errors but somewhat natural    & 0  & 0  & 2  & 0  & 0  \\
46--60: Grammatical but not natural          & 3  & 6  & 8  & 4  & 3  \\
61--75: Grammatical and somewhat natural     & \textbf{18} & \textbf{24} & \textbf{18} & \textbf{12} & \textbf{13} \\
76--90: Grammatical and natural (minor flaws)& 9  & 0  & 2  & 4  & 4  \\
91--100: Grammatical and Natural             & 0  & 0  & 0  & 0  & 0  \\
\midrule

% --- (c) DIALOGUE LEVEL: INFORMALITY ---
\multicolumn{6}{l}{\textbf{(c) Dialogue-Level: Informality}} \\
\midrule
00--15: Highly Formal                        & 0  & 1  & 0  & 0  & 0  \\
16--30: Very Formal                          & 0  & 8  & 0  & 3  & 0  \\
31--45: Moderately Formal                    & 1  & \textbf{10} & 1  & \textbf{5}  & 0  \\
46--60: Neutral                              & 8  & 7  & 6  & 8  & 0  \\
61--75: Slightly Informal                    & \textbf{20} & 4  & \textbf{22} & 4  & \textbf{14} \\
76--90: Very Informal                        & 1  & 0  & 1  & 0  & 6  \\
91--100: Extremely Informal / Casual         & 0  & 0  & 0  & 0  & 0  \\
\midrule

% --- (d) DIALOGUE LEVEL: COHESION & COHERENCE ---
\multicolumn{6}{l}{\textbf{(d) Dialogue-Level: Cohesion \& Coherence}} \\
\midrule
00--15: Incoherent and Incohesive            & 0  & 0  & 0  & 0  & 0  \\
16--30: Somewhat Coherent but Incohesive     & 0  & 0  & 0  & 0  & 0  \\
31--45: Somewhat coherent and cohesive       & 0  & 0  & 0  & 0  & 0  \\
46--60: Coherent and partially cohesive      & 2  & 2  & 3  & 2  & 3  \\
61--75: Somewhat coherent and cohesive       & 8  & \textbf{17} & \textbf{18} & \textbf{12} & \textbf{11} \\
76--90: Strong Coherence and Cohesion        & \textbf{20} & 11 & 9  & 6  & 6  \\
91--100: Fully Coherent and Cohesive         & 0  & 0  & 0  & 0  & 0  \\
\midrule

\textbf{Total Dialogues Evaluated}           & \textbf{30} & \textbf{30} & \textbf{30} & \textbf{20} & \textbf{20} \\
\bottomrule
\end{tabular}
\caption{\small SQM score distributions across language directions for \textit{Sarvam-Translate} fine-tuned on \textit{BaatCheet\_Human\_Train}, disaggregated by (a) Utterance-level Accuracy and Dialogue-level factors: (b) Naturalness, (c) Informality, and (d) Cohesion \& Coherence. Bold values indicate peak distribution bands.}
\label{tab:sarvam_sqm_distribution_combined}
\end{table*}

\section{Inter-Judge Correlation Analysis: Per-Model Breakdown}
\label{sec:appendix_correlation}

\paragraph{Metrics.}
We report \textbf{Pearson $r$} (linear magnitude agreement between 
continuous SQM scores) and \textbf{Kendall's $\tau$} (rank-order 
concordance across all pairwise dialogue comparisons). Together they 
provide complementary views: $r$ captures score-level agreement while 
$\tau$ captures ordinal consistency under non-normal distributions.

\paragraph{Utterance-Level.}
The strongest inter-judge agreement is observed at the utterance level 
(AVG: GPT--Human $r$=0.702, Gemini--Human $r$=0.697), confirming that 
sentence-level translation accuracy provides a more objective and 
consistently assessable criterion than holistic dialogue-level factors.

\paragraph{Naturalness and Cohesion \& Coherence.}
Both factors achieve moderate-to-good correlations (Naturalness AVG 
$r$: 0.55--0.74; Cohesion AVG $r$: 0.56--0.74 across judge pairs). 
Gemini--Human correlations consistently exceed GPT--Human across both 
dimensions, indicating that Gemini's evaluation tendencies better 
approximate human quality perception for informal Indic dialogue.

\paragraph{Informality.}
Informality yields the weakest agreement across all models and judge 
pairs (AVG $r$$<$0.40), with Kendall's $\tau$ values as low as 0.17. 
This reflects the inherent subjectivity of register assessment across 
code-mixed Dravidian languages, where dialectal variation, honorific 
calibration, and pragmatic code-switching resist standardization within 
a fixed rubric.

\paragraph{Overall Patterns.}
Three consistent trends emerge across all factors: 
(i)~GPT--Gemini yields the highest correlations, reflecting shared LLM 
tendencies in rubric interpretation; 
(ii)~Gemini--Human consistently exceeds GPT--Human, suggesting Gemini 
is a closer proxy for human judgement in this setting; and 
(iii)~Informality is universally the most disagreed-upon dimension, 
confirming it as the hardest qualitative factor to standardize across 
evaluators in multilingual informal dialogue assessment.

\begin{table*}[th!]
\centering
\scriptsize
\resizebox{\textwidth}{!}{%
\begin{tabular}{ll cc cc cc}
\toprule
& & \multicolumn{2}{c}{\textbf{GPT vs Human}} 
  & \multicolumn{2}{c}{\textbf{Gemini vs Human}} 
  & \multicolumn{2}{c}{\textbf{GPT vs Gemini}} \\
\cmidrule(lr){3-4} \cmidrule(lr){5-6} \cmidrule(lr){7-8}
\textbf{Factor} & \textbf{Model} & $r$ & $\tau$ & $r$ & $\tau$ & $r$ & $\tau$ \\
\midrule

\multirow{6}{*}{Naturalness}
& Gemma3-4B    & 0.544 & 0.361 & 0.574 & 0.395 & 0.687 & 0.504 \\
& Llama-3.2-3B & 0.540 & 0.382 & 0.606 & 0.427 & 0.758 & 0.548 \\
& Llama-3.1-8B & 0.594 & 0.422 & 0.665 & 0.486 & 0.788 & 0.605 \\
& Qwen3-4B     & 0.543 & 0.321 & 0.639 & 0.397 & 0.734 & 0.514 \\
& Sarvam-T     & 0.539 & 0.367 & 0.639 & 0.435 & 0.723 & 0.529 \\
& \textbf{AVG} & \textbf{0.552} & \textbf{0.370} & \textbf{0.625} & \textbf{0.428} & \textbf{0.738} & \textbf{0.540} \\

\addlinespace[4pt]
\multirow{6}{*}{Informality}
& Gemma3-4B    & 0.283 & 0.195 & 0.273 & 0.167 & 0.340 & 0.241 \\
& Llama-3.2-3B & 0.385 & 0.256 & 0.377 & 0.264 & 0.447 & 0.302 \\
& Llama-3.1-8B & 0.356 & 0.259 & 0.308 & 0.195 & 0.331 & 0.232 \\
& Qwen3-4B     & 0.347 & 0.232 & 0.320 & 0.188 & 0.373 & 0.236 \\
& Sarvam-T     & 0.344 & 0.217 & 0.383 & 0.231 & 0.466 & 0.321 \\
& \textbf{AVG} & \textbf{0.343} & \textbf{0.232} & \textbf{0.332} & \textbf{0.209} & \textbf{0.391} & \textbf{0.266} \\

\addlinespace[4pt]
\multirow{6}{*}{\shortstack{Cohesion \&\\Coherence}}
& Gemma3-4B    & 0.468 & 0.323 & 0.558 & 0.398 & 0.624 & 0.440 \\
& Llama-3.2-3B & 0.651 & 0.462 & 0.695 & 0.503 & 0.869 & 0.684 \\
& Llama-3.1-8B & 0.535 & 0.377 & 0.648 & 0.471 & 0.738 & 0.546 \\
& Qwen3-4B     & 0.605 & 0.318 & 0.694 & 0.431 & 0.787 & 0.501 \\
& Sarvam-T     & 0.547 & 0.342 & 0.638 & 0.408 & 0.699 & 0.490 \\
& \textbf{AVG} & \textbf{0.561} & \textbf{0.364} & \textbf{0.647} & \textbf{0.442} & \textbf{0.743} & \textbf{0.532} \\

\addlinespace[4pt]
\multirow{6}{*}{\shortstack{Utterance\\Level}}
& Gemma3-4B    & 0.714 & 0.517 & 0.706 & 0.541 & 0.773 & 0.595 \\
& Llama-3.2-3B & 0.818 & 0.622 & 0.806 & 0.624 & 0.878 & 0.706 \\
& Llama-3.1-8B & 0.682 & 0.507 & 0.679 & 0.539 & 0.788 & 0.613 \\
& Qwen3-4B     & 0.665 & 0.477 & 0.662 & 0.503 & 0.765 & 0.584 \\
& Sarvam-T     & 0.632 & 0.367 & 0.630 & 0.420 & 0.704 & 0.485 \\
& \textbf{AVG} & \textbf{0.702} & \textbf{0.498} & \textbf{0.697} & \textbf{0.525} & \textbf{0.781} & \textbf{0.597} \\

\bottomrule
\end{tabular}%
}
\caption{\small Per-model inter-judge correlation ($r$ = Pearson, $\tau$ = Kendall's $\tau$) across all four evaluation factors between GPT-4o-mini, Gemini-Flash-Lite, and Human Expert Linguists. Bold rows indicate cross-model averages.}
\label{tab:appendix_inter_judge_full}
\end{table*}

\end{document}